\documentclass[11pt]{article}

\usepackage[a4paper,margin=1in]{geometry}
\usepackage[T1]{fontenc}
\usepackage[utf8]{inputenc}
\usepackage{lmodern}
\usepackage{microtype}
\usepackage{booktabs}
\usepackage{array}
\usepackage{longtable}
\usepackage{tabularx}
\usepackage{enumitem}
\usepackage{hyperref}
\usepackage{url}
\usepackage{xcolor}
\usepackage{tikz}
\usetikzlibrary{arrows.meta,positioning,shapes.geometric,fit,calc}
\usepackage{listings}
\usepackage{caption}
\usepackage{amsmath}
\usepackage{amssymb}
\usepackage{graphicx}
\graphicspath{{figures/}}

\hypersetup{
  colorlinks=true,
  linkcolor=blue!50!black,
  citecolor=blue!50!black,
  urlcolor=blue!50!black,
  pdftitle={CHAP: A Technical Guide to the Collaborative Human-Agent Protocol},
}

\definecolor{chaporange}{HTML}{EA4700}
\definecolor{chaplight}{HTML}{FFF8F2}
\definecolor{chapagent}{HTML}{FFE7DD}
\definecolor{chaptool}{HTML}{E8F1ED}
\definecolor{chappeer}{HTML}{EDE0F2}
\definecolor{codebg}{HTML}{F7F7F8}

\lstdefinestyle{chapjson}{
  basicstyle=\ttfamily\small,
  backgroundcolor=\color{codebg},
  frame=single,
  rulecolor=\color{black!15},
  breaklines=true,
  columns=fullflexible,
  showstringspaces=false,
  numbers=left,
  numberstyle=\tiny\color{black!45},
  xleftmargin=1.8em,
  framexleftmargin=1.5em,
  extendedchars=true,
  inputencoding=utf8,
  literate=
    {§}{{\S}}1
    {£}{{\pounds}}1
    {Δ}{{$\Delta$}}1
    {—}{{---}}1
    {–}{{--}}1
    {'}{{'}}1
    {'}{{'}}1
    {"}{{``}}1
    {"}{{''}}1
}
\newcommand{\chap}{\textsc{chap}}

\title{\vspace{-1.2cm}\textbf{Collaborative Human-Agent Protocol (CHAP)}\\
\large An open protocol for auditable, structured multi-human and multi-agent collaboration}
\author{Arsalan Shahid, Gordon Suttie, and Philip Black\\Brightbeam AI\\ \\ \texttt{arsalan.shahid@brightbeam.com}}
\date{June 2026}

\begin{document}
\maketitle

\begin{abstract}

Foundation models are moving from response generation into operational roles. They plan across steps, call tools, request human input, coordinate with other agents, and increasingly carry responsibility for work that affects customers, claims, code, contracts, and clinical decisions. Production deployments are no longer one human supervising one model. They are multi-human, multi-agent collaborations that cross teams, time zones, and trust boundaries.
The technical surface for this collaboration remains weakly specified. When an agent drafts a response and a human edits it before it ships, the moment of human judgement is the most valuable signal in the system. In current practice it is recorded, if at all, in application code, chat threads, ticket comments, and tribal memory. Two protocol standards address adjacent concerns: MCP standardises agent access to tools and data, and A2A standardises agent-to-agent interoperability. Neither defines the shared workspace in which humans and agents perform accountable work together.
This paper presents CHAP, the Collaborative Human-Agent Protocol. Under CHAP, the override that used to vanish into a chat thread becomes a structured event carrying a diff, a rationale, and a content hash. The handoff between shifts becomes a portable envelope rather than a pinned message. The human approval of an agent's draft becomes a non-repudiable signed decision that can be replayed years later. The protocol achieves this through a small Core (workspaces, participants, tasks, artefacts, and an append-only evidence log) together with composable profiles that add review, modes, routing, deliberation, handoff, identity, signatures, and transparency-backed audit as deployments require them. Specification, two interoperable reference implementations (TypeScript and Python), conformance harness, and worked examples are available at: \url{https://github.com/BrightbeamAI/chap}
\end{abstract}


\pagebreak
\tableofcontents
\newpage

\section{Introduction}

What must be standardised when humans and AI agents share accountable work?

The first generation of foundation-model applications treated the model as an assistant: a user asked a question, the model produced an answer, and the interaction ended. That pattern remains useful, but it is no longer sufficient. In operational settings, AI systems increasingly plan across steps, call tools, consult records, ask for clarification, coordinate with other agents, and produce outputs that humans must review, modify, approve, or reject. The technical question has therefore shifted. The central problem is not only what a single model can generate. It is how humans and agents work together around a shared problem, under explicit policy, with evidence that can be replayed later.

This paper calls that target state the \emph{Human--Agent Symphony}: a mode of work in which humans, agents, and services participate in a shared operational space. The term is aspirational, but the requirement is practical. A production system needs to know who delegated a task, who accepted it, what evidence was used, what the agent produced, whether a human reviewed it, how the human changed it, why an escalation occurred, and what final action was taken. These are not merely user-interface details. They are collaboration events. In governed environments, they must be represented, exchanged, stored, and audited.

\begin{figure}[!ht]
\centering
\includegraphics[width=\textwidth]{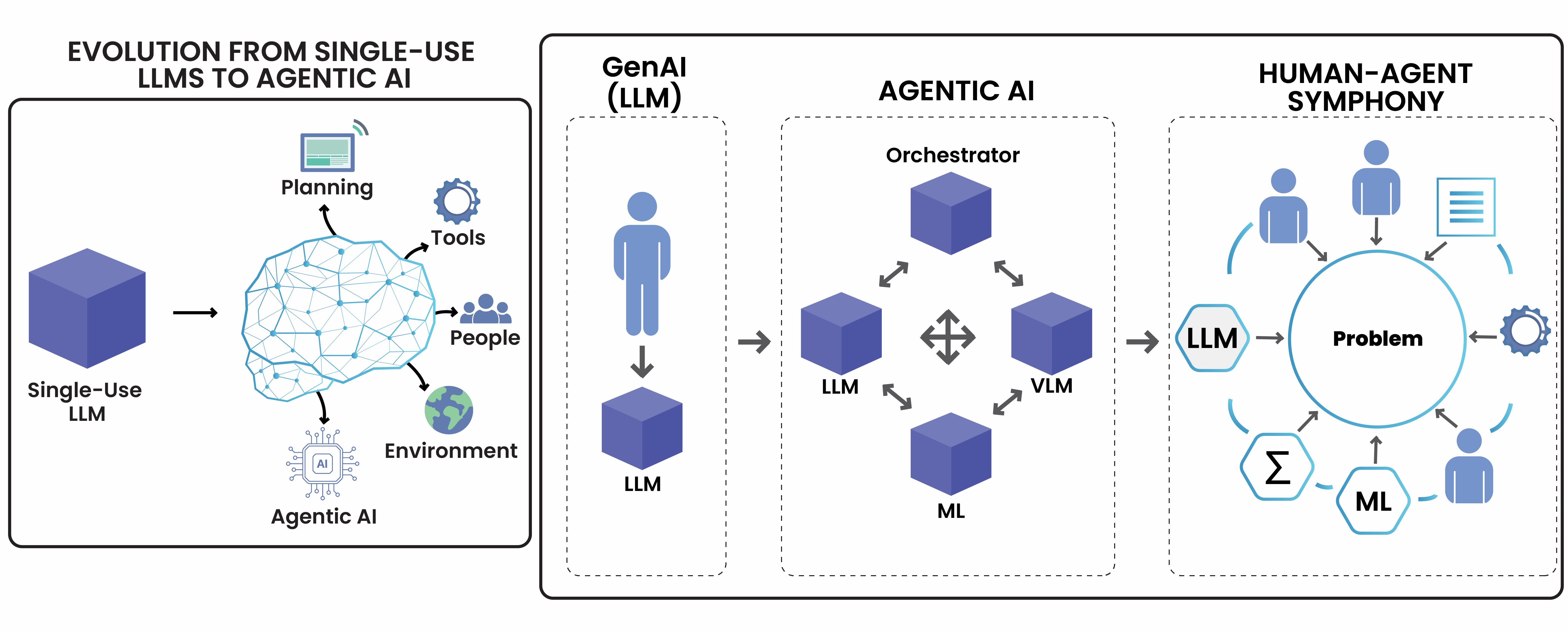}
\caption{Three waves in the evolution of agentic systems. Wave~I centred on isolated conversational assistants. Wave~II added planning, memory, tool use, and early multi-agent orchestration. Wave~III centres on shared human-agent workspaces where humans, agents, and services collaborate under explicit policy and shared audit.}
\label{fig:evolution}
\end{figure}

Figure~\ref{fig:evolution} summarises the progression from isolated assistants to shared human-agent workspaces.

\textbf{Wave~I: assistant interaction.} Early foundation-model applications normalised a simple pattern: one user, one model, one session. The system generated text, code, images, summaries, or recommendations, but the collaboration surface remained narrow. The user interface carried most of the interaction semantics.

\textbf{Wave~II: agentic execution.} The next wave added planning, tool use, memory, and agent-to-agent coordination. Agents could decompose work, call external systems, consult data sources, and delegate subtasks. Protocols such as the Model Context Protocol (MCP) and Agent2Agent (A2A) began to define important boundaries in this ecosystem: MCP for agent access to tools and data, and A2A for interoperability between independently developed agents.

\textbf{Wave~III: shared accountable work.} The current challenge is broader. The unit of work is no longer a single model response or a single agent turn. It is a problem that may involve several humans, several agents, existing enterprise systems, policy constraints, review checkpoints, and audit requirements. This wave requires a collaboration layer that is visible to the protocol itself.

Consider a support triage workflow. A supervisor assigns a refund case to a triage agent. The agent retrieves order history, checks customer records, drafts a response, and proposes a discretionary credit. The proposed credit exceeds a policy threshold, so the agent requests human review. A reviewer edits the wording, changes the credit amount, and records the reason as a goodwill adjustment. Because the revised credit exceeds a second threshold, the case moves into deliberation with another reviewer. Once approved, another agent issues the credit through a tool call. Weeks later, an auditor needs to replay the case: the assignment, evidence, draft, review, human edit, rationale, escalation, approval, tool invocation, and final outcome.

Most enterprise teams can build some version of this flow today. They usually do so by encoding the collaboration semantics inside product-specific application code. That approach works locally, but it fragments the system boundary. Review decisions look different in a ticketing tool, a document platform, and an agent orchestration framework. Human edits may be saved only as final text, not as structured diffs. Abstention may be indistinguishable from timeout. Shift handoff may live in chat rather than in the task state machine. Audit logs may show that something happened without showing why a human approved, rejected, or modified an agent's output.

\chap{}, the Collaborative Human--Agent Protocol, addresses this gap. It defines a shared workspace in which humans, agents, services, groups, and bridge participants are protocol-visible actors. It standardises recurring collaboration events such as task assignment, acceptance, progress, completion, review request, approval, rejection, override, abstention, escalation, whisper, handoff, deliberation, pause, resume, snapshot, and rollback. Each accepted message becomes part of an append-only evidence log. Each human override can be represented as a typed diff with rationale and tags. Each escalation or abstention becomes a first-class event rather than an implementation-specific side effect.

\chap{} does not replace agent frameworks, workflow engines, tool protocols, identity providers, policy engines, or audit stores. It composes with them. Its purpose is narrower: to define the shared room where accountable human-agent work happens, and to define what counts as evidence inside that room.

\subsection{The protocol gap}

The agent ecosystem has begun to standardise several adjacent boundaries. MCP defines a portable way for an agent or host application to reach tools, prompts, resources, and data sources. A2A defines a portable way for agents and agent platforms to discover each other and exchange work. Workflow engines such as Temporal, Airflow, and Argo provide durable execution and scheduling. Identity systems, policy engines, and transparency logs provide authentication, authorisation, governance, and audit infrastructure.

None of these layers, on its own, defines the collaboration grammar for a shared human-agent workspace. A workflow engine can schedule a task, but it does not define what a structured human override looks like. An identity provider can authenticate a participant, but it does not define how that participant approves, rejects, abstains, or escalates. A tool protocol can expose a data source, but it does not define how the resulting evidence is reviewed by a human. An agent-to-agent protocol can connect two agents, but it does not define the accountable workspace in which humans and agents jointly carry work to completion.

\chap{} fills that missing layer. A \chap{} workspace has participants, active profiles, policy references, an operational mode, tasks, artefacts, and an evidence log. Participants may be humans, agents, services, groups, or bridge participants representing work in another system. The protocol does not assume that all participants have equal authority. Rather, it makes each participant visible, typed, and accountable under the workspace policy. The result is a portable collaboration surface that can be implemented across different products, runtimes, and organisational settings.

\subsection{Status of this document}

This document is a v0.2 working draft. Sections describing Core data structures, envelope fields, method names, lifecycle states, profile behaviour, and conformance levels are intended as specification material. User journeys, runtime algorithms, deployment patterns, implementation guidance, and appendices are informative unless explicitly marked otherwise.

\paragraph{Maturity.} \chap{} v0.2 is a public \emph{Draft}. The protocol surface, schemas, and reference implementations are stable enough for experimentation and early production pilots; they are not yet sufficient for a normative conformance claim under a standards-track convention. The specification ships with two interoperable reference implementations: the TypeScript \texttt{@chap/coordinator} package and the Python \texttt{chap-coordinator} package, both in the public repository at \url{https://github.com/BrightbeamAI/chap}~\cite{chap-repo}. Both implement every method in the v0.2 catalogue and pass the same conformance harness against the published test vectors on the same JSON-RPC 2.0 wire. An empirical interoperability test suite covering the full method surface is published as canonical test vectors and a runnable harness; cross-implementation negative tests and profile-specific fixtures continue to mature. Breaking changes to the wire format follow Semantic Versioning, but the profile surface should be expected to evolve faster than Core. Production deployments are welcome and encouraged to feed back findings; deployments requiring stability guarantees beyond reasonable best effort under SemVer should wait for 1.0.

\subsection{Contributions and structure of this report}

This report makes three main contributions. First, it identifies the shared human-agent workspace as a distinct interoperability problem, separate from agent-to-tool access, agent-to-agent messaging, workflow scheduling, and identity management. Second, it specifies the \chap{} Core: workspaces, participants, tasks, artefacts, evidence entries, a task lifecycle, and a JSON-RPC-2.0-inspired envelope for exchanging collaboration events. Third, it defines a progressive profile model for review, modes, routing, whisper channels, handoff, deliberation, control, identity, signatures, and SCITT-backed audit, supported by implementation guidance, worked examples, conformance levels, and two interoperable v0.2 reference implementations.

The rest of the report is organised as follows. Section~\ref{sec:background} situates \chap{} in relation to existing agent protocols, workflow systems, identity infrastructure, and audit mechanisms. Section~\ref{sec:chap-glance} gives a compact overview of Core and profiles. Section~\ref{sec:architecture} defines the protocol architecture, workspace boundary, envelope model, lifecycle, artefacts, and evidence chain. Sections~\ref{sec:data-model} and~\ref{sec:methods-profiles} define the data model, method surface, and profile catalogue. Sections~\ref{sec:runtime-semantics},~\ref{sec:security}, and~\ref{sec:conformance} describe runtime semantics, assurance considerations, and conformance. Sections~\ref{sec:implementation},~\ref{sec:deployment-patterns},~\ref{sec:standards}, and~\ref{sec:user-journeys} provide implementation guidance, deployment patterns, standards positioning, and informative journeys. The report closes with limitations, future work, appendices, and references.

\section{Background and Need}
\label{sec:background}

\subsection{From assistants to operational participants}

An assistant responds to a prompt. An operational agent participates in work. The difference is not only model capability; it is system structure. An operational agent must operate inside a contract that defines what work it may perform, who delegated that work, what evidence it used, when it must ask for help, how its output is reviewed, and what happens when a human modifies or rejects its recommendation. Prompt design, model selection, and orchestration frameworks can improve agent behaviour, but they do not by themselves define this collaboration contract. The challenge sits at the long-recognised intersection of mixed-initiative interfaces~\cite{horvitz1999}, calibrated reliance on automation~\cite{lee2004}, and human-AI interaction design~\cite{amershi2019}; what is missing is a portable wire-level vocabulary for the collaboration events those literatures describe.

The requirement is clearest in regulated and operationally critical environments such as healthcare, biopharma, insurance, financial services, public administration, advanced manufacturing, cybersecurity, and support operations. In these settings, the question is not only whether a model can generate a plausible answer. The organisation must also be able to show who requested the work, who or what performed it, which policy applied, what evidence was considered, whether a human reviewed or changed the result, and how the system behaved when uncertainty, risk, or authority limits were reached. Each of these facts corresponds to a collaboration event. Each event needs a stable representation on the wire.

\subsection{Why existing protocols are not enough}

\chap{} is designed to compose with existing standards and infrastructure rather than replace them. JSON-RPC 2.0~\cite{jsonrpc} provides a familiar request-response pattern that informs the \chap{} envelope. MCP~\cite{mcp} addresses agent access to tools, resources, prompts, and external data. A2A~\cite{a2a} addresses interoperability between independently developed agents and agent platforms. OpenID Connect~\cite{oidc} and OAuth 2.0~\cite{oauth} support authentication and authorisation. W3C Verifiable Credentials~\cite{vc20} can carry machine-verifiable claims about participants. SCITT~\cite{scitt} provides an architecture for transparency-backed supply-chain and audit evidence. Workflow engines provide durable execution, retries, scheduling, and state management.

These layers are necessary, but none of them defines the shared human-agent workspace. A tool protocol can expose a customer record, but it does not define how that record becomes evidence in a human review. An agent-to-agent protocol can delegate work between agents, but it does not define how a human supervisor approves, rejects, overrides, or escalates the resulting task. An identity provider can authenticate a user or service, but it does not define the lifecycle of accountable collaboration. A workflow engine can schedule a step, but it does not define the meaning of a structured override, abstention, deliberation, or handoff.

Table~\ref{tab:protocol-positioning} draws the boundary.

\begin{table}[h]
\centering
\caption{Protocol positioning. \chap{} composes with adjacent protocols rather than competing with them.}
\label{tab:protocol-positioning}
\begin{tabularx}{\textwidth}{@{}p{2.6cm}X X@{}}
\toprule
\textbf{Layer} & \textbf{Primary concern} & \textbf{Typical boundary} \\
\midrule
MCP & Agent access to tools, prompts, resources, and external data. & Agent or host application $\leftrightarrow$ tool/resource server. \\
A2A & Interoperability between independently developed agents and agent platforms. & Agent $\leftrightarrow$ agent, often across systems or organisations. \\
\chap{} & Shared workspace semantics: task lifecycle, human review, override, abstention, escalation, handoff, deliberation, evidence, and audit. & Human $\leftrightarrow$ agent $\leftrightarrow$ service inside an auditable workspace. \\
\bottomrule
\end{tabularx}
\end{table}

\begin{figure}[!ht]
\centering
\includegraphics[width=0.8\textwidth]{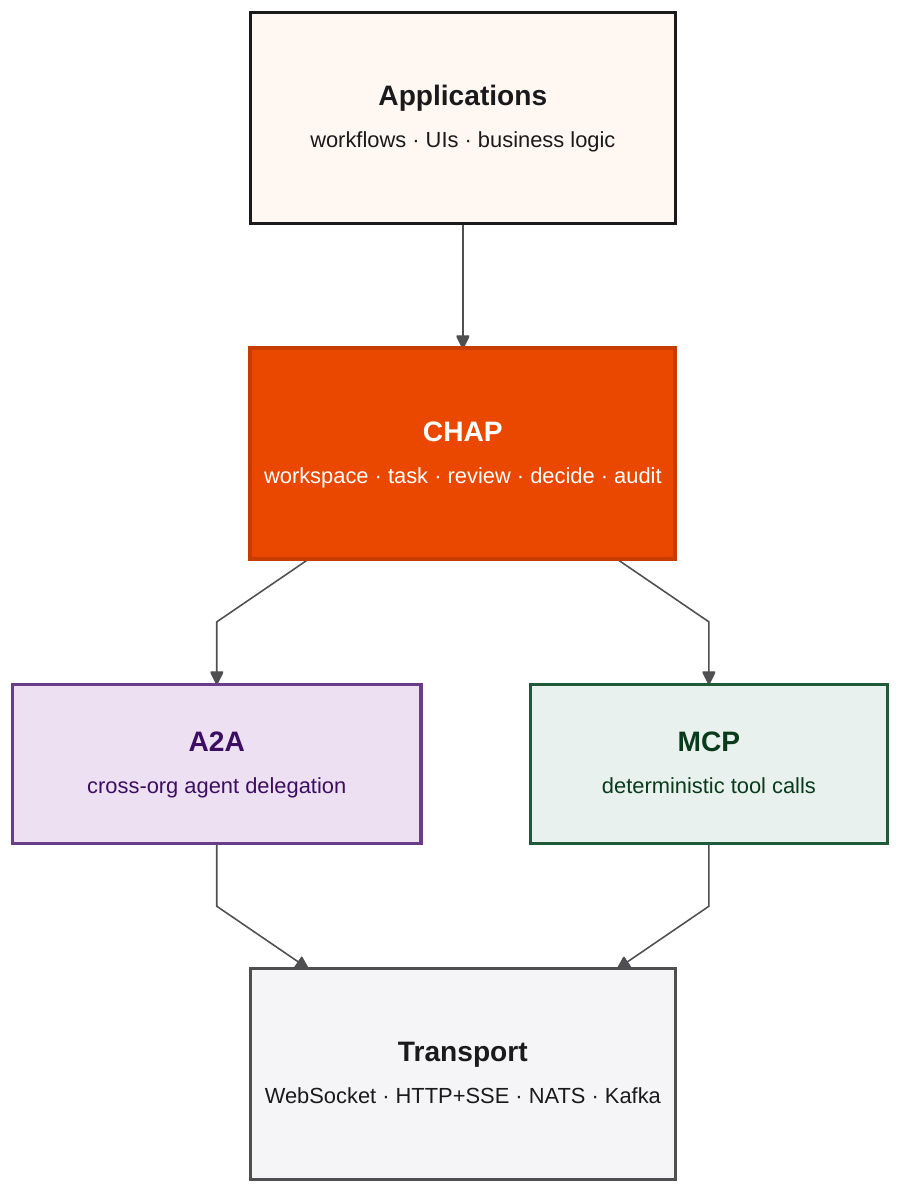}
\caption{The agent protocol stack. MCP handles agent-to-tool and agent-to-resource access; A2A handles agent-to-agent interoperability; \chap{} handles the shared human-agent workspace and its evidence trail. Identity, policy, workflow, and transparency systems plug into this stack rather than being replaced by it.}
\label{fig:stack}
\end{figure}

The three layers describe different boundaries in an agentic system. MCP connects agents to the tools and data they need. A2A connects agents to other agents and agent platforms. \chap{} defines the collaboration space in which humans, agents, and services carry accountable work to completion. Figure~\ref{fig:stack} shows how these layers fit with identity, audit, policy, and workflow infrastructure.

\subsection{Requirements for governed human-agent work}

\chap{} is motivated by nine recurring requirements in governed human-agent deployments.

\begin{description}[leftmargin=!,labelwidth=1.2cm]
    \item[R1] \textbf{Shared workspace.} Humans, agents, services, groups, and bridge participants need a common namespace for tasks, artefacts, membership, policies, modes, and evidence.

    \item[R2] \textbf{Typed task lifecycle.} Assignment, acceptance, progress, completion, review, abstention, escalation, cancellation, and supersession should be protocol-visible events rather than user-interface conventions.

    \item[R3] \textbf{Structured human review.} Human approval, rejection, modification, and request-for-changes should be represented in a reusable format rather than left as unstructured comments.

    \item[R4] \textbf{Structured override.} A human edit to an agent output should be recordable as a typed diff with rationale and tags, so it can support audit, workflow improvement, policy tuning, and agent evaluation.

    \item[R5] \textbf{Explicit abstention and escalation.} A principled ``I should not decide this'' must be distinguishable from silence, timeout, failure, or low-confidence generation.

    \item[R6] \textbf{Operational modes.} Shadow, trial, and production modes should be visible at the protocol level so that new agents can be evaluated before their outputs affect the real world.

    \item[R7] \textbf{Composability.} Tool calls, peer-agent delegation, identity, policy checks, signatures, and transparency audit should reuse existing infrastructure where possible.

    \item[R8] \textbf{Portable evidence.} The evidence chain should be replayable and verifiable without depending on a single user interface, vendor, or runtime.

    \item[R9] \textbf{Incremental adoption.} A minimal \chap{} deployment should be useful without requiring every optional profile, identity binding, or cryptographic audit mechanism.
\end{description}

\subsection{What \chap{} deliberately does not define}
\label{sec:non-goals}

The protocol's usefulness depends on being narrow. \chap{} addresses the collaboration layer between humans, agents, and services in a shared workspace; it deliberately does not address the following concerns, which belong in profiles, in deploying applications, or in the policy and regulatory layer above the protocol.

\begin{description}[leftmargin=!,labelwidth=1.2cm]
    \item[N1] \textbf{A claim or evidence taxonomy.} \chap{} carries an artefact's content and citations opaquely. Whether claims are typed as Evidence / Inference / Assumption / Gap / Recommendation, as Premise / Conclusion, or as some domain-specific scheme is a profile contribution or a deploying-application choice, not a Core obligation.

    \item[N2] \textbf{A temporal model beyond \texttt{produced\_at} and chain monotonicity.} Domains that require richer time semantics, separating subject time from statement time, or carrying validity-from / validity-until windows, should layer those into the artefact content shape or define them in a profile. Core preserves only the minimum needed to keep the evidence chain ordered and replayable.

    \item[N3] \textbf{A confidence calibration.} The \texttt{routing\_hints.confidence} field on a task or artefact is a model-reported number. \chap{} makes no claim about cross-model comparability or about what any particular value implies for routing or review depth; calibration is the operator's responsibility.

    \item[N4] \textbf{What evidence is sufficient for any regulatory regime.} \chap{} produces a verifiable record of who decided what, when, and on the basis of which inputs. Whether that record meets a particular audit, conformity assessment, or accountability standard (the EU AI Act~\cite{euai} conformity assessment, the NIST AI Risk Management Framework~\cite{nistairmf}, GMP Annex 11, SOX-404 control reliance, IAF/SEAR personal accountability, DORA, or any other) is for the deploying organisation and its regulators to determine. \chap{} provides infrastructure; sufficiency is a socio-technical determination.

    \item[N5] \textbf{Semantic relations between artefacts beyond \texttt{based\_on} and supersession.} Richer graphs (causation, mitigation, verification, intervention) belong in domain extraction layers above \chap{}. The protocol preserves enough structural relation (override, supersession, citation) to allow such a graph to be derived externally without forcing its vocabulary into the wire format.
\end{description}

These boundaries matter because the alternative is a protocol that tries to be a universal grammar for collaborative work and ends up either bloated, sector-specific, or both. \chap{}'s posture is to surface the small set of events common across domains and to let richer vocabularies sit above it in clearly versioned profiles.

\section{CHAP at a Glance}
\label{sec:chap-glance}

\subsection{Core plus profiles}

\chap{} is organised as a minimal Core with optional profiles. Core defines the primitives needed for accountable collaboration: workspaces, participants, coordinators, tasks, artefacts, evidence entries, and the message envelope. Profiles add capabilities when a workflow requires them. This design supports progressive adoption. A small internal deployment can begin with Core. A regulated workflow can add Review, Modes, Identity, Signing, and Audit-SCITT. A cross-organisational deployment can add routing, bridge participants, verifiable identity claims, and stronger audit anchoring.

\begin{figure}[!ht]
\centering
\begin{tikzpicture}[font=\small, node distance=0.7cm]
  \tikzstyle{box}=[draw, rounded corners, align=center, text width=0.92\textwidth, inner sep=8pt]
  \node[box, fill=chaplight] (profiles) {\textbf{Profiles}\\ security-signed $\cdot$ audit-scitt $\cdot$ identity-oidc $\cdot$ identity-vc $\cdot$ review $\cdot$ whisper\\ deliberation $\cdot$ modes $\cdot$ routing $\cdot$ handoff $\cdot$ control};
  \node[box, fill=chapagent, below=of profiles] (core) {\textbf{Core}\\ workspace $\cdot$ participant $\cdot$ coordinator $\cdot$ task $\cdot$ artefact $\cdot$ evidence $\cdot$ message envelope};
\end{tikzpicture}
\caption{\chap{} adoption model. Core is useful on its own. Profiles are layered when a workspace needs additional collaboration, identity, control, security, or audit capability.}
\label{fig:core-profiles}
\end{figure}
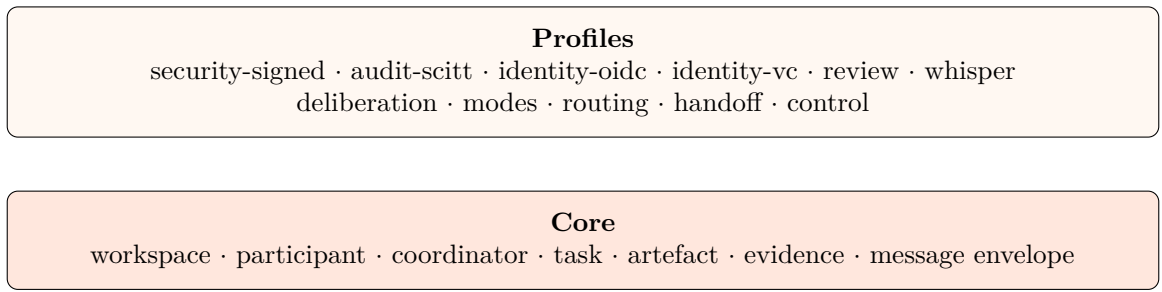

\subsection{Core primitives}

The Core model is built around six primary primitives.

\begin{longtable}{@{}p{3.1cm}p{11.3cm}@{}}
\caption{Core \chap{} primitives.}
\label{tab:core-primitives}\\
\toprule
\textbf{Primitive} & \textbf{Meaning} \\
\midrule
\endfirsthead
\toprule
\textbf{Primitive} & \textbf{Meaning} \\
\midrule
\endhead
Workspace & A bounded collaboration space with an identifier, state, roster, active profiles, policy references, operational mode, tasks, artefacts, and evidence log. \\
Participant & Any protocol-visible entity that can send or receive \chap{} messages, including a human, agent, service, group, or bridge participant. \\
Coordinator & A service participant that receives envelopes, checks workspace policy and mode constraints, routes accepted messages, and appends accepted envelopes to the evidence log. \\
Task & A unit of work that can be proposed, assigned, accepted, started, progressed, reviewed, completed, escalated, cancelled, or superseded inside a workspace. \\
Artefact & A typed record associated with a task, such as a draft, decision, review outcome, override diff, route decision, citation set, handoff summary, or deliberation result. \\
Evidence entry & An append-only record of an accepted envelope, optionally hash-linked, signed, and externally anchored. \\
\bottomrule
\end{longtable}

\begin{figure}[!ht]
\centering
\includegraphics[width=0.62\textwidth]{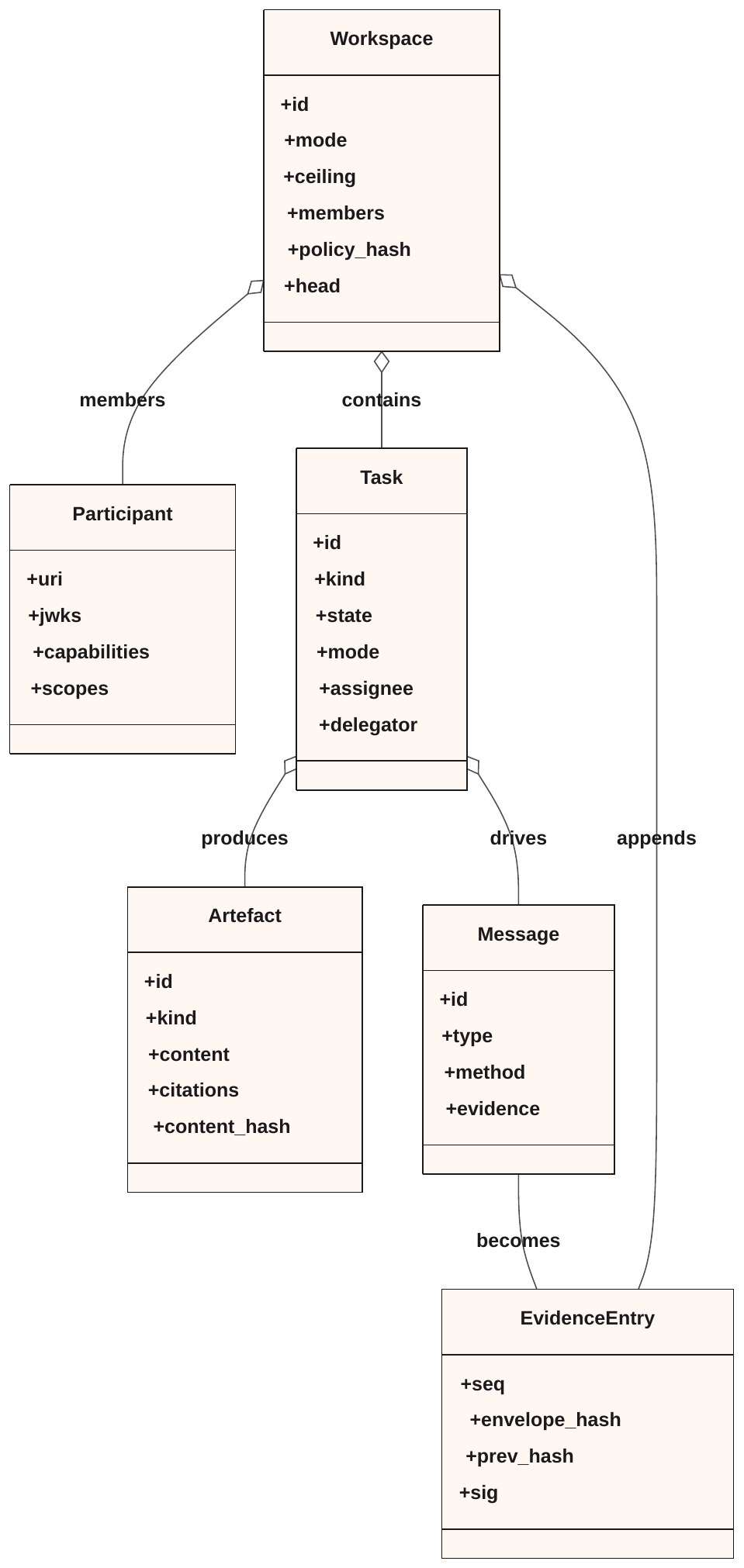}
\caption{Conceptual relations among \chap{} primitives. A workspace has participants, active profiles, policy references, an operational mode, and a stream of tasks. Tasks produce artefacts. Accepted envelopes become evidence entries. The diagram is generated from the same Mermaid source as the reference implementation documentation.}
\label{fig:primitives}
\end{figure}

\section{Protocol Architecture}
\label{sec:architecture}

\subsection{The workspace as the collaboration boundary}

A workspace is the unit of collaboration and audit in \chap{}. It scopes participant membership, active profiles, task identifiers, routing rules, policy references, operational mode, and evidence-chain state. A workspace can represent a small internal flow, such as a human reviewer and a drafting agent, or a larger operational setting, such as a multi-shift support desk, a regulated approval board, or a federated incident-response environment.

A workspace is not a chat room, although chat-like interfaces can be built on top of it. It is also not a workflow engine. Its purpose is to define and record collaboration semantics: who participated, what work was assigned, what evidence was used, what artefact was produced, what decision was taken, and how that decision changed the state of the work. Durable execution, scheduling, retries, queues, and long-running business processes can remain in existing workflow infrastructure. \chap{} records the human-agent task and decision events that matter for interoperability, governance, and audit.

\subsection{Participants and identity}

A participant is any protocol-visible entity that can send or receive \chap{} messages. Participants are represented by URI-like identifiers with a type prefix. Common examples include \texttt{human:alice@example.org}, \texttt{agent:triage-bot\#v3.2}, \texttt{service:coordinator@example.org}, 

\noindent \texttt{group:review-board@example.org}, and \texttt{workspace:wsp\_incident\_response}. A participant descriptor records the participant type, role, declared capabilities, scopes, identity bindings, and keys where applicable.

\chap{} Core does not mandate a specific identity provider. This is intentional. Many deployments already have enterprise identity, access-management, and audit systems. The \texttt{identity-oidc} profile binds human participants to OpenID Connect tokens and OAuth scopes. The \texttt{identity-vc} profile binds participants to W3C Verifiable Credentials when richer cross-organisational claims are required, such as professional role, regulated authorisation, delegated authority, or external accreditation. The \texttt{security-signed} profile binds participant keys to message signatures.

\subsection{The Coordinator}

The Coordinator is a service participant that mediates the workspace. It receives envelopes, validates message shape, checks participant authority, applies workspace policy and mode constraints, routes accepted messages, applies profile-specific rules, and appends accepted messages to the evidence log. A simple deployment can run a single Coordinator instance. A production deployment can run Coordinator replicas behind a load balancer, with a transactional database, event store, or append-only log as the state backend.

The Coordinator is not a trust-free oracle. In minimal deployments, it is a trusted service. Stronger deployments can make the Coordinator more accountable by combining participant signatures, hash-linked evidence entries, transparency receipts, and external anchoring. These mechanisms do not remove the need to operate the Coordinator securely, but they reduce the ability of a compromised or faulty Coordinator to alter history without detection.

\subsection{Envelope model}

\chap{} envelopes use a JSON-RPC-2.0-inspired request, response, and notification shape with \chap{}-specific extension fields. The method and parameter structure follows the JSON-RPC convention; the additional fields identify the protocol version, workspace, origin, recipient, timestamp, message type, and evidence metadata. Implementations should reject envelopes that are not valid for the advertised Core version and active profiles.

A representative request envelope is shown in Listing~\ref{lst:envelope}. The example is illustrative; profile-specific fields may add signatures, identity claims, routing metadata, or transparency receipts.

\begin{lstlisting}[caption={Representative \chap{} request envelope.},label={lst:envelope}]
{
  "jsonrpc": "2.0",
  "chap": "0.2",
  "id": "01HZ9YWQ7K3X8M2V4N6P8R0T2A",
  "ts": "2026-05-17T09:14:22.184Z",
  "workspace": "wsp_support_triage",
  "from": "human:alice@example.org",
  "to": "agent:triage-bot#v3.2",
  "type": "request",
  "method": "task.assign",
  "params": {
    "kind": "draft_customer_response",
    "input": {
      "ticket_id": "INC-48910",
      "customer_message": "Order arrived broken; please refund."
    },
    "routing_hints": {
      "criticality": "medium",
      "deadline": "2026-05-17T10:00:00Z",
      "risk_tier": "customer_retention"
    }
  },
  "evidence": {
    "prev_hash": "sha256:5f1c...ccbb",
    "sig": "ed25519:k-2026-05-17a:V8M2...q0kg=="
  }
}
\end{lstlisting}

\subsection{Task lifecycle}

The task lifecycle is intentionally small. It captures enough state to support delegation, execution, review, escalation, completion, and audit without turning \chap{} into a general-purpose workflow engine. Workflow systems may maintain richer internal state, but the \chap{} lifecycle defines the state transitions that need to be visible across participants and replayable through the evidence log.

\begin{figure}[!ht]
\centering
\includegraphics[width=\textwidth]{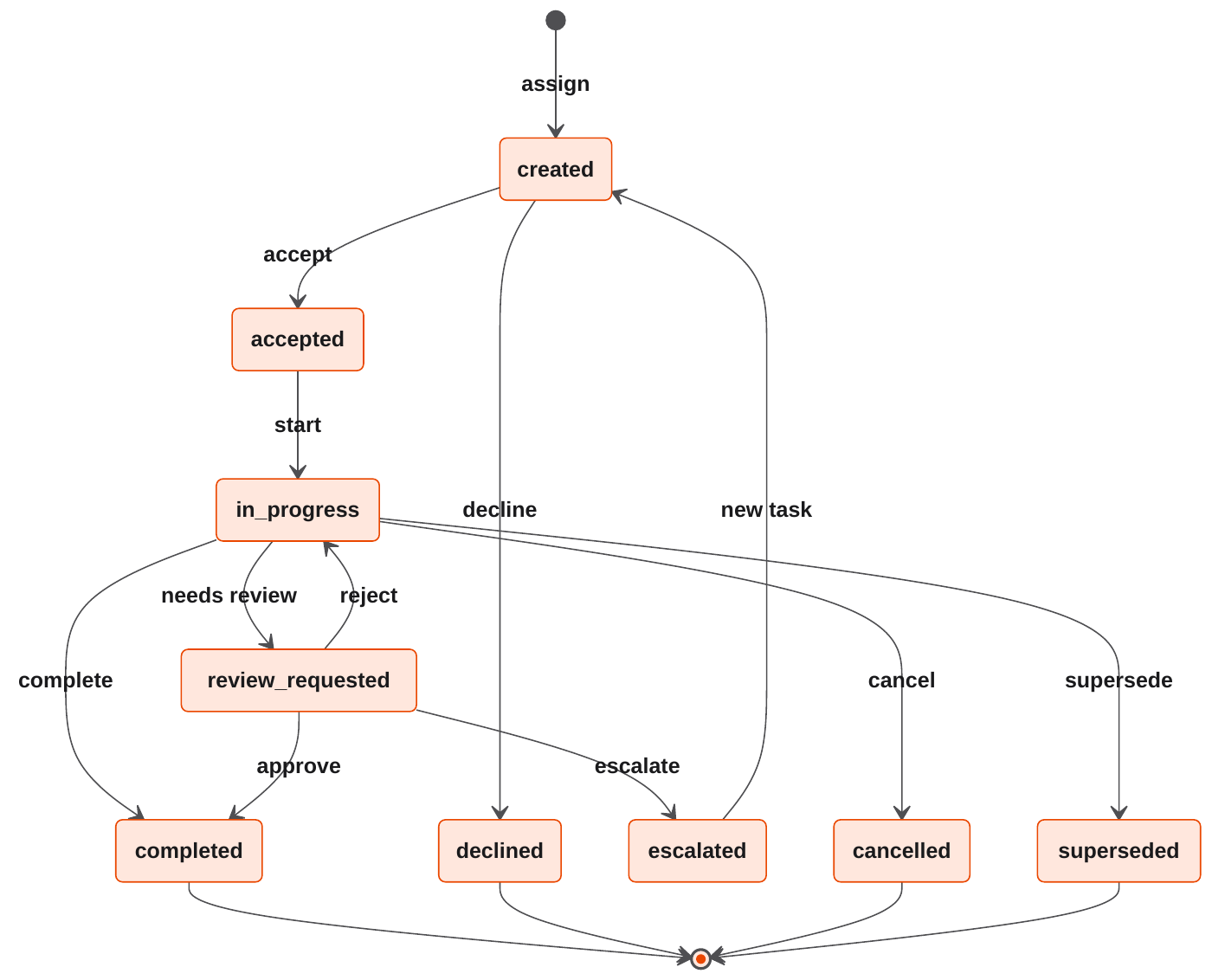}
\caption{Simplified \chap{} task lifecycle. Core transitions capture the basic movement of accountable work. Profiles add specialised transitions, such as review request, deliberation, and mode promotion, without changing the role of the evidence log. Every accepted transition becomes an evidence entry.}
\label{fig:task-lifecycle}
\end{figure}

\subsection{Artefacts and citations}

A task produces artefacts. An artefact can be a draft answer, decision, structured record, route decision, handoff summary, override, deliberation result, or domain-specific payload declared by an implementation. Artefacts may carry citations to external evidence. When an agent uses MCP to call tools, \chap{} should not duplicate the full MCP transcript. Instead, the resulting artefact can cite the relevant tool call, transcript, or result, including hashes of inputs and outputs where available. This preserves a verifiable audit boundary while keeping \chap{} focused on the collaboration layer. The same pattern applies to A2A exchanges with remote peers: the bridge participant cites the A2A correlation by hash rather than absorbing the remote traffic into the local chain. The v0.2 reference implementations ship library helpers (\texttt{wrap\_mcp\_tool\_call}, \texttt{wrap\_a2a\_message\_exchange}) that emit this citation pattern from either language. The helpers are convenience over the bare protocol; the pattern itself is a property of the data model, not the helpers.

\subsection{Evidence chain}

Every accepted envelope becomes an evidence entry. In minimal Core deployments, the evidence chain may be an ordered append-only audit log. In signed deployments, entries can be hash-linked and signed by the originating participant, the Coordinator, or both, depending on the active profile and deployment policy. In the \texttt{audit-scitt} profile, evidence entries can be represented as SCITT statements and associated with transparency receipts that a third party can verify.

\begin{figure}[!ht]
\centering
\includegraphics[width=0.7\textwidth]{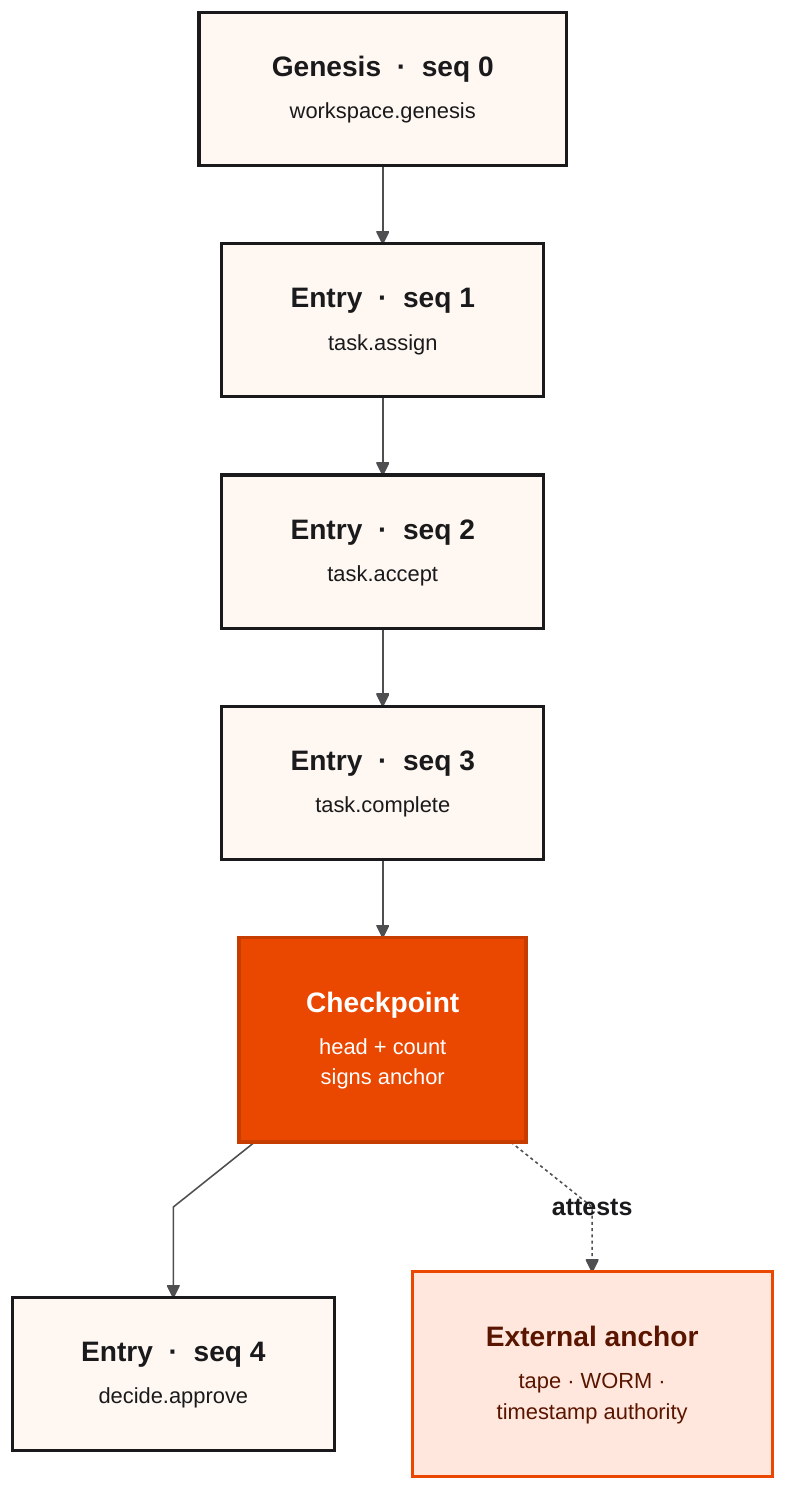}
\caption{Evidence chain. Each accepted envelope can reference the previous entry's hash. In Core deployments, this can be an ordered append-only log. In signed deployments, entries are signed according to workspace policy. In deployments using the \texttt{audit-scitt} profile, entries can be anchored as SCITT signed statements with externally verifiable transparency receipts.}
\label{fig:evidence}
\end{figure}

\section{Data Model}
\label{sec:data-model}

The \chap{} data model defines the records that participants and Coordinators exchange, store, project, and audit. This section summarises the main records. The machine-readable schemas in the repository provide the validation source for implementers; the examples below are representative rather than exhaustive.

\subsection{Workspace descriptor}

A workspace descriptor records the state of a bounded collaboration space. It identifies the workspace, advertised profiles, operational mode, membership, policy references, and current evidence head. Implementations may add deployment-specific metadata, but they should preserve the core fields needed for interoperability and replay.

\begin{lstlisting}[caption={Representative workspace descriptor.},label={lst:workspace}]
{
  "id": "wsp_support_triage",
  "state": "active",
  "profiles": ["core/1.0", "review/1.0", "modes/1.0", "routing/1.0"],
  "mode": "trial",
  "mode_ceiling": "production",
  "policy_uri": "policy://support-triage/v4",
  "members": [
    {"uri": "human:alice@example.org", "type": "human", "role": "reviewer"},
    {"uri": "agent:triage-bot#v3.2", "type": "agent", "role": "drafter"},
    {"uri": "service:coordinator@example.org", "type": "service", "role": "coordinator"}
  ],
  "evidence_head": "sha256:9d7b..."
}
\end{lstlisting}

The descriptor returned by \texttt{workspace.describe} should also expose enough information for participants to determine which methods, profiles, modes, and policy references are active. A participant should not assume support for a profile-specific method unless the workspace advertises the relevant profile.

\subsection{Participant descriptor}

A participant descriptor records a participant URI, type, role, capabilities, scopes, identity bindings, keys, and status. Participant types include humans, agents, services, groups, and bridge participants. Agents should disclose capabilities, model identifiers, version identifiers, tool access, and operating constraints where practical. Human participants should bind to identity profiles when the deployment requires verified accountability, privileged decisions, or cross-organisational trust.

The descriptor is not intended to expose secrets or unnecessary personal data. It should expose only the information needed for routing, authorisation, accountability, and audit.

\subsection{Task descriptor}

A task descriptor represents a unit of work inside a workspace. It includes an identifier, kind, state, delegator, assignee, input, output references, artefacts, deadline, routing hints, review metadata, supersession references, and route decisions. A task may also expose a history view for application convenience.

Task history is a projection, not the source of truth. The evidence log remains the authoritative record of accepted envelopes and state transitions. Applications may cache task state for performance, but they should be able to reconstruct the protocol-visible history from evidence entries.

\subsection{Artefact descriptor}

An artefact descriptor represents a typed output or intermediate record associated with a task. Artefacts may include drafts, decisions, review outcomes, override diffs, route decisions, citation sets, handoff summaries, deliberation results, or domain-specific payloads. A descriptor should identify the producing participant, artefact kind, content or content URI, content hash, citations, policy references, routing hints, and relevant timestamps.

\paragraph{Artefact identity: \texttt{id}, \texttt{logical\_id}, \texttt{instance\_id}.} \chap{} distinguishes three identity concepts on an artefact. The \texttt{id} field (required) is a globally unique handle for this particular artefact record; each new artefact gets a fresh identifier. The optional \texttt{logical\_id} field names the durable thing the artefact is about, the handle that survives revision, override, and supersession. Two artefacts that share a \texttt{logical\_id} are two versions of the same underlying item: the same draft response, the same policy statement, the same recommendation. Producers should assign a \texttt{logical\_id} on first creation and reuse it on every subsequent revision. The optional \texttt{instance\_id} field is a stable handle for the specific version; when present, an \texttt{instance\_id} must equal the artefact's \texttt{content\_hash} or be deterministically derived from it, so consumers can detect whether two artefacts with the same \texttt{logical\_id} are byte-identical.

These fields exist so that revision, supersession, and override can be distinguished in the evidence chain. Without them, a deployment can track \emph{which artefact replaced which} (via \texttt{based\_on} and \texttt{control.supersede}) but cannot easily answer the question \emph{``is this the same item I approved last week, or a different item with the same shape?''} That question arises in any domain that does versioned work. \chap{} itself reads only \texttt{id}; higher layers (analytics, dashboards, external indexes) can use \texttt{logical\_id} and \texttt{instance\_id} to project the chain into a version graph.

Sensitive artefacts should not be copied into immutable evidence entries unless necessary. A deployment may store sensitive content externally, encrypt it, and reference it from \chap{} using a URI, content hash, retention policy, and access-control metadata.

\section{Method Surface and Profile Catalogue}
\label{sec:methods-profiles}

\subsection{Core methods}

The \chap{} Core method surface is designed to be implementable in a short engineering cycle. The v0.2 reference path exposes a compact set of seven methods: \texttt{workspace.describe}, \texttt{participant.join}, \texttt{participant.leave}, \texttt{task.create}, \texttt{task.update}, \texttt{task.complete}, and \texttt{audit.read}. This compact path is sufficient for a minimal conformant deployment.

The protocol vocabulary also defines more specific lifecycle verbs, including \texttt{task.assign}, \texttt{task.accept}, \texttt{task.decline}, \texttt{task.start}, and \texttt{task.progress}. A compact implementation may represent these through \texttt{task.create} and \texttt{task.update}; a richer implementation may expose them as explicit methods. Implementations should advertise the supported method surface through \texttt{workspace.describe}.

\begin{table}[h]
\centering
\caption{Practical Core method set in the v0.2 reference path.}
\label{tab:core-methods}
\begin{tabularx}{\textwidth}{@{}p{4.3cm}X@{}}
\toprule
\textbf{Method} & \textbf{Purpose} \\
\midrule
\texttt{workspace.describe} & Return workspace descriptor, state, participants, tasks, policy references, modes, and advertised profiles. \\
\texttt{participant.join} & Add a human, agent, service, group, or workspace participant to the active workspace membership. \\
\texttt{participant.leave} & Remove a participant from the active workspace membership. \\
\texttt{task.create} & Create and dispatch a task in the compact reference path. \\
\texttt{task.update} & Transition a task's state or record progress in the compact reference path. \\
\texttt{task.complete} & Submit a completed task output and optional artefact metadata. \\
\texttt{audit.read} & Read an ordered range of evidence entries from the workspace evidence log. \\
\bottomrule
\end{tabularx}
\end{table}

\subsection{Profiles}

Profiles add optional capability while keeping Core small. A workspace advertises the profiles it supports. Participants should not assume profile-specific methods, artefact types, or validation rules unless the relevant profile is active in the workspace.

\begin{longtable}{@{}p{4.2cm}p{10cm}@{}}
\caption{\chap{} profile catalogue.}
\label{tab:profiles}\\
\toprule
\textbf{Profile} & \textbf{Capability} \\
\midrule
\endfirsthead
\toprule
\textbf{Profile} & \textbf{Capability} \\
\midrule
\endhead
\texttt{review/1.0} & Request review, approve, reject, override, abstain, and escalate. This is the highest-value early profile because it turns human review and human edits into structured collaboration data. \\
\texttt{modes/1.0} & Add \texttt{shadow}, \texttt{trial}, and \texttt{production} modes so agents can be evaluated and promoted under explicit constraints. \\
\texttt{routing/1.0} & Interpret task and artefact routing hints to choose assignees, review depth, and escalation behaviour. \\
\texttt{whisper/1.0} & Support narrow, deadline-bound questions during task execution, with typed options and default behaviour if the answer lapses. \\
\texttt{handoff/1.0} & Transfer active work between participants, including shift changes, follow-the-sun operations, and graceful reassignment. \\
\texttt{deliberation/1.0} & Support multi-human decision rules such as quorum, all-approve, weighted vote, and weighted vote with veto. \\
\texttt{control/1.0} & Pause, resume, cancel, supersede, snapshot, roll back, and set mode ceilings. \\
\texttt{security-signed/1.0} & Add canonicalised message signatures, key references, and signature verification rules. \\
\texttt{audit-scitt/1.0} & Represent selected evidence entries as SCITT transparency-backed signed statements and receipts. \\
\texttt{identity-oidc/1.0} & Bind human participants to OIDC tokens, OAuth scopes, token-to-key binding, and step-up authentication. \\
\texttt{identity-vc/1.0} & Bind participants to W3C Verifiable Credentials for richer claims and cross-organisational identity. \\
\bottomrule
\end{longtable}

\subsection{Review and override}

The Review profile is central to \chap{} because many production deployments require human approval, rejection, modification, or escalation before an agent output can affect the real world. The profile defines explicit review requests and review decisions. Its most distinctive event is \texttt{decide.override}: a human approves a modified version of an agent artefact and records the change as an override artefact.

An override artefact should preserve the base snapshot, the structured diff, the resulting artefact, the reviewer, the rationale, tags, policy references, and timestamp. When the artefact being overridden carries a \texttt{logical\_id} (Section~\ref{sec:data-model}), the override should carry the same \texttt{logical\_id} and should set the optional \texttt{intent\_preserved} flag. The flag distinguishes \emph{the human refined the expression of the same underlying decision} (\texttt{true}) from \emph{the human substituted a different decision} (\texttt{false}). The field is informational; \chap{} does not constrain semantics. It exists because ``the human edited the agent's draft'' and ``the human replaced the agent's draft with a different decision'' are operationally different events that produce identical envelope structures without it. The same convention applies to \texttt{control.supersede}: when superseding an artefact that carries a \texttt{logical\_id}, the replacement should carry the same \texttt{logical\_id} and set \texttt{intent\_preserved} accordingly.

Listing~\ref{lst:override} shows a representative example.

\begin{lstlisting}[caption={Representative override artefact.},label={lst:override}]
{
  "id": "art_override_01HZ...",
  "task_id": "tsk_01HZ...",
  "reviewer": "human:alice@example.org",
  "based_on_artefact": {
    "kind": "draft",
    "content_hash": "sha256:3ce1..."
  },
  "logical_id": "lgl_01HZ...",
  "intent_preserved": true,
  "diff": [
    {
      "op": "replace",
      "path": "/body/paragraphs/1",
      "value": "I can absolutely see why this is frustrating."
    },
    {
      "op": "add",
      "path": "/body/paragraphs/3",
      "value": "I have also requested a goodwill credit for your account."
    }
  ],
  "result": {
    "subject": "Re: damaged order",
    "body": "...final approved response..."
  },
  "rationale": "Tone was too procedural for an eight-year customer with repeated failures.",
  "tags": ["tone-warmed", "goodwill-credit-suggested"],
  "policy_refs": ["support.refunds.v4", "goodwill-credit.limit"],
  "ts": "2026-05-17T09:23:40Z"
}
\end{lstlisting}

This record supports four downstream uses. It is an audit record because it shows what changed and why. It is a supervision signal because the human provides a corrected output rather than a binary approval or rejection. It is a policy signal because repeated overrides may reveal unclear thresholds or missing rules. It is an adoption signal because persistent override patterns show where agent behaviour diverges from human judgement. The \texttt{intent\_preserved} flag refines the supervision signal: a high rate of refining overrides indicates the agent reaches the right decision in the wrong way (a delivery problem), while a high rate of substituting overrides indicates the agent reaches the wrong decision (a judgement problem). These tune to different fixes.

\subsection{Modes and promotion}

The Modes profile defines a promotion ladder. In \texttt{shadow} mode, an agent's output is recorded but not delivered. In \texttt{trial} mode, output can be delivered but every result is reviewed. In \texttt{production} mode, output can be delivered according to workspace policy, often with sampling, routing, confidence thresholds, and high-risk exceptions.

\begin{figure}[!ht]
\centering
\includegraphics[width=0.45\textwidth]{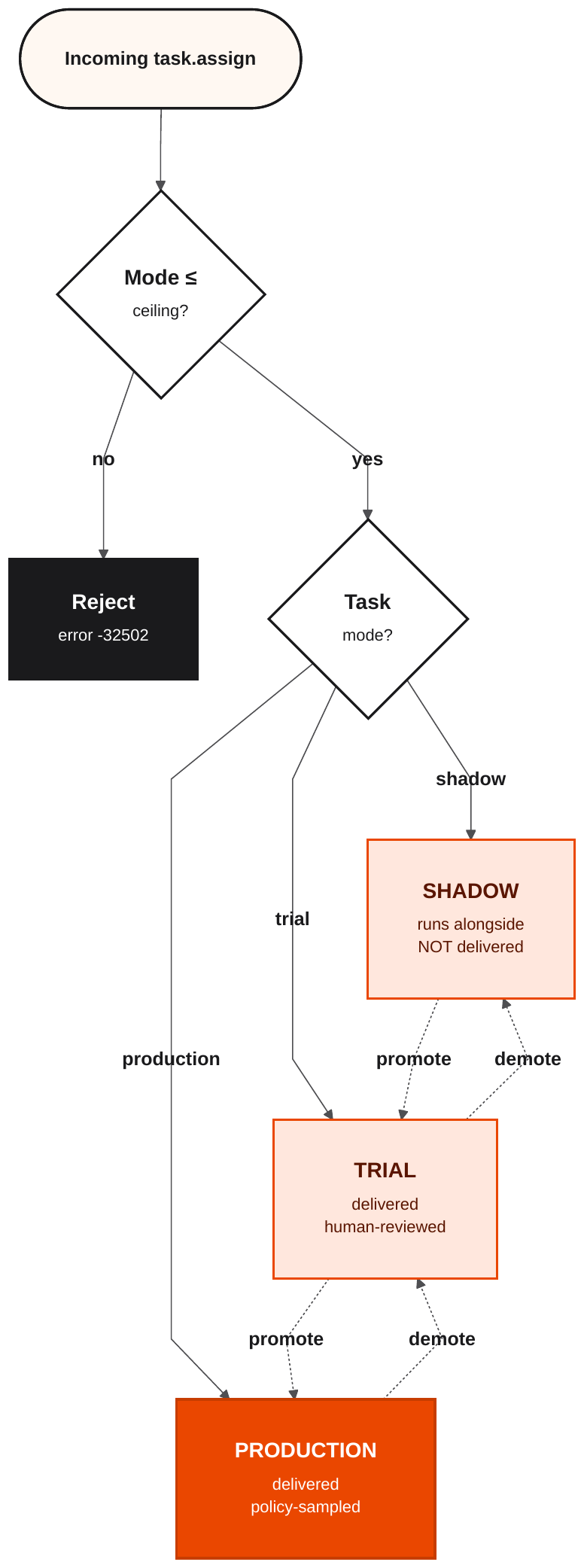}
\caption{Mode promotion ladder. An agent starts in \emph{shadow}: its output is recorded but never delivered. In \emph{trial}, output can be delivered, but every result is reviewed. In \emph{production}, the workspace policy decides what gets reviewed and what does not. Promotion is gated by evidence the operator collects from the chain: agreement rate, override rate, abstention rate, latency, cost.}
\label{fig:modes}
\end{figure}

Recommended promotion evidence includes agreement rate with human decisions, override rate, abstention rate, incident rate, latency, cost, task coverage, policy violations, and performance stability across shifts, teams, and data conditions. \chap{} does not prescribe thresholds because thresholds are domain-specific; it ensures that the evidence needed for those thresholds is recorded consistently.

\subsection{Routing}

Core carries routing signals; the Routing profile defines decisions that consume those signals. For example, a task may include \texttt{criticality}, \texttt{deadline}, \texttt{max\_cost\_usd}, and \texttt{risk\_tier}. An artefact may include \texttt{confidence}, \texttt{model\_id}, \texttt{cost\_consumed\_usd}, and \texttt{latency\_ms}. The Routing profile then records decisions such as selecting an assignee, choosing review depth, or auto-escalating.

This split is important. The protocol carries evidence; the operator runs policy. A regulated insurer and a support desk may both use \chap{} signals, but they should not be forced into the same routing policy.

\subsection{Whisper}

Whisper prompts support narrow, deadline-bound interrupts during work. A whisper is not a review because no completed artefact exists yet. It is not chat because the answer is typed, bounded, and defaulted. For example, an agent may ask whether to refund or replace, which order to cancel when ambiguity exists, or whether to use a conservative interpretation of a policy clause.

\subsection{Handoff and deliberation}

Handoff supports the transfer of active work between participants. It is useful for shift changes, escalation, follow-the-sun operations, or participant unavailability. Deliberation supports multi-party decisions with explicit rules: any-one-approves, all-approve, quorum, weighted vote, and weighted vote with veto. Both are common in human organisations and therefore need protocol-level representation when agents participate in the same work.

\subsection{Control}

The Control profile provides operational commands: pause, resume, cancel, supersede, snapshot, rollback, and set mode ceiling. These commands are privileged and audit-visible. Rollback appends corrective evidence rather than truncating history, preserving the audit chain.

\subsection{Integration packages: CHAP as MCP server and A2A agent}
\label{sec:integration-packages}

The protocol composes with MCP and A2A in two directions. The outward direction, described in Section~\ref{sec:standards} and journeys 8 and 9, lets a CHAP workspace cite external tool calls and bridge to remote agents. The inward direction lets external MCP clients and A2A orchestrators drive a CHAP workspace through the protocol they already speak.

The v0.2 repository ships transport adapters in both languages. The MCP adapter (\texttt{@chap/coordinator-mcp} for TypeScript, \texttt{chap\_coordinator.transports.mcp\_server} for Python) wraps a Coordinator as an MCP server and exposes every CHAP method as a tool named \texttt{chap.<method>}. Any MCP client, including Claude Desktop, Cursor, and Claude Code, can list those tools and call them. The A2A adapter (\texttt{@chap/coordinator-a2a} for TypeScript, \texttt{chap\_coordinator.transports.a2a\_server} for Python) does the same for A2A: every CHAP method becomes an \texttt{AgentSkill} on the published Agent Card. A2A-aware orchestrators register the coordinator by URL and delegate work to it.

Both adapter layers are thin. They translate inbound MCP tool calls or A2A messages into JSON-RPC envelopes and dispatch through the Coordinator. The adapter holds no state. The audit chain a workspace produces is byte-identical whether the envelopes arrived through the inward MCP transport, the inward A2A transport, or directly over JSON-RPC. This means the inward adapters are a transport binding, not a wire-format change, and Section~\ref{sec:conformance} therefore treats them as deployment options rather than conformance levels.

The two A2A SDK ecosystems are at different points of the A2A spec evolution. The TypeScript SDK targets A2A 0.3.0 and the Python SDK targets A2A 1.0. The CHAP adapter layer is identical across the two. Agent Cards advertise the correct version per implementation. The MCP adapter ecosystem is more uniform: both the TypeScript and Python adapters target MCP 2025-11-25.

Runnable reference servers ship for each combination: stdio servers for MCP at \texttt{reference/mcp-server-ts/} and \texttt{reference/mcp-server-py/}, HTTP servers for A2A at \texttt{reference/a2a-server-ts/} (Express) and \texttt{reference/a2a-server-py/} (FastAPI). Five-minute walkthroughs in the repository describe how to wire each into an external client.

\section{Runtime Semantics and Informative Algorithms}
\label{sec:runtime-semantics}

The algorithms in this section are informative. They illustrate runtime behaviour that conformant implementations should preserve, while allowing different engineering choices for storage, transport, deployment, and policy execution.

\subsection{Envelope acceptance}

A Coordinator should apply a consistent acceptance pipeline. The exact implementation may vary, but the logical sequence should preserve validation, authorisation, policy enforcement, atomic state transition, and evidence append.

\begin{lstlisting}[caption={Coordinator acceptance pipeline, informative pseudocode.},label={lst:acceptance}]
function accept(envelope):
    parse envelope as JSON
    validate JSON-RPC / CHAP envelope shape
    verify workspace exists and is active
    verify method is known in Core or an active profile
    verify from participant is known, unless method permits join
    verify recipient is valid for the method and workspace
    verify participant role, scope, and authority for the method
    verify freshness, replay window, and idempotency constraints

    if security-signed enabled:
        canonicalize envelope excluding signature material
        verify participant signature and key validity

    apply method-specific transition rules
    enforce mode ceiling and workspace policy constraints

    atomically:
        persist state transition
        append evidence entry

    publish response or notification to recipients
\end{lstlisting}

The crucial property is atomicity. A workspace should not move to a new protocol-visible state without an evidence entry, and an evidence entry should not exist for a transition that was rejected.

\subsection{Review depth decision}

The Routing profile can decide review depth using task hints, artefact metadata, workspace mode, and deployment policy. \chap{} does not prescribe the policy. It specifies how the resulting decision can become evidence.

\begin{lstlisting}[caption={Example review-depth policy, deployment-specific.},label={lst:reviewdepth}]
function review_depth(task, artefact, workspace):
    if workspace.mode == "shadow":
        return "record_only"

    if workspace.mode == "trial":
        return "full_review"

    if task.routing_hints.criticality in ["high", "critical"]:
        return "full_review"

    if artefact.routing_hints.confidence < 0.70:
        return "full_review"

    if artefact.routing_hints.confidence < 0.85:
        return "spot_check"

    return "skip_or_sample"
\end{lstlisting}

The output of this function should be recorded as a route-decision artefact containing the observed hints, policy identifier, outcome, and rationale.

\subsection{Mode promotion}

Mode promotion should be governed by evidence rather than enthusiasm. A practical promotion assessment can aggregate task counts, override rates, rejection rates, abstention rates, escalation rates, policy violations, latency, cost, and incident reports. Human sign-off remains central. \chap{} can store the evidence and the mode transition; it cannot decide the acceptable risk threshold for every domain.

\begin{lstlisting}[caption={Evidence-based mode promotion logic, informative.},label={lst:promotion}]
function may_promote(agent_version, from_mode, to_mode, window):
    metrics = aggregate_chap_evidence(agent_version, window)

    require metrics.task_count >= policy.minimum_tasks
    require metrics.critical_incidents == 0
    require metrics.policy_violations == 0
    require metrics.override_rate <= policy.max_override_rate[to_mode]
    require metrics.rejection_rate <= policy.max_rejection_rate[to_mode]
    require metrics.abstention_rate within policy.expected_abstention_band
    require human_signoff(policy.required_roles)

    return true
\end{lstlisting}

\subsection{Override analytics}

Structured overrides can be analysed without scraping final documents. Useful aggregations include overrides by model version, task kind, policy reference, reviewer role, artefact kind, route decision, and tag. This analysis can support prompt improvement, retrieval improvement, policy clarification, reviewer training, workflow redesign, and carefully governed fine-tuning datasets.

Override records should not be treated as automatic ground truth. Human reviewers can also be inconsistent, biased, rushed, or constrained by local practice. \chap{} captures the signal in a structured form. Governance determines whether and how that signal is used for model improvement, process redesign, or policy change.

\section{Security, Trust, and Compliance}
\label{sec:security}

\subsection{Threat model}

\chap{} addresses security-relevant collaboration concerns: message integrity, participant accountability, policy enforcement, evidence-chain integrity, and mode safety. The threat classes below name the adversary, the protocol-level countermeasure, and (where the protocol cannot defend alone) the deployment-level mitigation. The full operational threat model is maintained in the public \texttt{SECURITY.md} document; this section names the surface that is normative to the specification.

\begin{description}[leftmargin=!,labelwidth=2.6cm]
    \item[\textbf{Replay.}] An adversary captures a previously-valid envelope and re-injects it. \emph{Countermeasures:} envelope identifiers are ULIDs that conformant Coordinators must reject on second observation; timestamps must be monotonically non-decreasing per originator; \texttt{prev\_hash} must match the current chain head, so any replay against an advanced chain fails acceptance.

    \item[\textbf{Downgrade.}] An adversary forces capability negotiation to advertise fewer profiles than both peers support, hoping to suppress a defensive profile such as \texttt{security-signed} or \texttt{audit-scitt}. \emph{Countermeasures:} the workspace descriptor is itself an artefact in the evidence chain; its advertised profile list is signed and cannot be retrospectively narrowed. Deployments concerned about downgrade should treat the profile set as policy and refuse any participant whose \texttt{participant.join} declares a lower profile set than the workspace's mandatory minimum.

    \item[\textbf{Capability confusion.}] Two profiles define methods with similar names but different security properties. \emph{Countermeasures:} methods are namespaced (\texttt{namespace.verb}); the profile that owns a namespace is declared in the workspace descriptor; Coordinators must reject method calls whose namespace's owning profile is not in the advertised set.

    \item[\textbf{Key rotation.}] A participant rotates a signing key mid-chain. \emph{Countermeasures:} \texttt{identity-oidc} and \texttt{identity-vc} define key rotation as an explicit \texttt{participant.update} event signed by the old key and naming the new key. Verifiers must treat post-rotation entries as signed by the new key only after the rotation event itself has been verified by the old key; rotation events must not retroactively re-sign earlier entries.

    \item[\textbf{Evidence-chain forking under partition.}] Two Coordinators serving the same workspace under a network partition each accept envelopes into their local chain head; on heal the chains have diverged. \emph{Countermeasures:} the evidence chain is per-workspace and per-Coordinator; \chap{} does not provide a Byzantine fault tolerant consensus layer. Deployments requiring continuity through partition must either run a single logical Coordinator with HA replication preserving chain linearity, or operate the peer-to-peer topology with each peer maintaining its own chain and using \texttt{audit.read} to cross-verify on heal. Fork detection is automatic (divergent \texttt{prev\_hash} values do not link); resolution is operational. Anchoring chain heads to an external transparency log via \texttt{audit-scitt} makes fork detection independent of the Coordinators themselves.

    \item[\textbf{Compromised Coordinator.}] An adversary controls a Coordinator and attempts to forge, suppress, or rewrite entries. \emph{Countermeasures:} signatures are made by originating participants, not by the Coordinator, so the Coordinator cannot forge new participant content. Suppression of a delivered envelope is detectable because the emitting participant retains a record. Rewriting history breaks \texttt{prev\_hash} linkage and, where deployed, breaks the SCITT receipt's witnessed root. A Coordinator that is the sole signer of receipts can equivocate; deployments defending against this must use \texttt{audit-scitt} with an externally operated transparency service whose witnesses are not under the same administrative control as the Coordinator.

    \item[\textbf{Identity confusion.}] A participant adopts a Participant URI that resembles another's. \emph{Countermeasures:} Participant URIs in \texttt{human:}, \texttt{agent:}, \texttt{service:} namespaces must be bound to a verified identity (OIDC subject claim or VC subject DID) before being admitted to a workspace via \texttt{participant.join}; the binding is recorded in the participant descriptor and signed.
\end{description}

\paragraph{Out of scope.} \chap{} does not defend against: a Participant who chooses to lie within the schema (a human who approves without reading the artefact; an agent that hallucinates a citation); the semantic content of artefacts (the protocol carries opaque content; semantic integrity is the deploying application's concern); side-channel inference on \texttt{routing\_hints} or other metadata; or denial-of-service at the transport layer (handled by the underlying transport's controls). \chap{} also does not by itself secure external tool servers, model weights, prompts, deployment infrastructure, network boundaries, or application secrets. A secure deployment must combine \chap{} with appropriate identity, access control, infrastructure security, monitoring, incident response, and data-protection controls.

\subsection{Signing and canonicalisation}

The \texttt{security-signed} profile uses deterministic canonicalisation and participant signatures. The intended verification path is: remove the signature material, canonicalise the envelope, verify the signature against the participant's active key at the relevant timestamp, check key revocation and rotation state, and verify evidence-chain linkage.

Ed25519 is used in the v0.2 profile because it is compact, fast, and widely implemented. Future versions may add additional signature suites, including hybrid post-quantum options, if deployment requirements justify them.

\subsection{Identity binding}

OIDC is the recommended path for ordinary enterprise human identity. A participant may sign \chap{} messages with a local key bound to an authenticated session through token-to-key binding, DPoP, or an equivalent mechanism. Step-up authentication can be required for privileged operations such as promotion to production, rollback, workspace mode changes, policy changes, or authority-sensitive approvals.

Verifiable Credentials are more appropriate when claims need to be issued by a party other than the participant's employer, when selective disclosure matters, or when regulated role claims need to travel across organisational boundaries.

\subsection{Audit and transparency}

A Core evidence log establishes operational traceability. A signed evidence chain strengthens tamper evidence. A SCITT-backed audit profile can provide transparency receipts and third-party-verifiable log inclusion. The design principle is to avoid inventing a new transparency primitive. \chap{} profiles existing transparency infrastructure for workspace evidence rather than replacing it.

\subsection{Privacy, retention, and redaction}

Evidence logs can conflict with privacy, confidentiality, and retention obligations if they store full sensitive content. \chap{} should therefore support opaque artefact references, content hashes, external encrypted storage, retention policies, access-control metadata, and redaction markers.

The v0.2 draft explicitly recognises that confidentiality for evidence-chain content is not fully solved at the protocol level. Production deployments should avoid placing unnecessary personal, sensitive, regulated, or commercially confidential content directly in immutable evidence entries. Where immutable records are required, deployments should prefer references, hashes, encrypted stores, and controlled disclosure over raw content duplication.

\subsection{Regulated AI governance}

For organisations subject to AI governance requirements, \chap{} can provide infrastructure for traceability, human oversight, post-deployment monitoring, accountability, and incident review. It does not make a system compliant by itself. Compliance depends on the use case, risk classification, data governance, model validation, human factors, organisational process, security controls, and legal obligations.

\chap{}'s role is to make operational evidence available in a structured and interoperable form. That evidence can support governance, but governance remains a socio-technical responsibility of the deploying organisation.

\section{Conformance and Evaluation}
\label{sec:conformance}

\subsection{Conformance levels}

The v0.2 conformance model defines two implementable levels (Minimal, Recommended) and a planned third (Full). An implementation may claim a level only against the method set it has actually implemented and exercised against the published test vectors; a claim against a method declared but not implemented is non-conformant. The levels are intended to support progressive adoption while making clear which capabilities are required for higher-assurance deployments.

\begin{table}[h]
\centering
\caption{v0.2 conformance levels.}
\label{tab:conformance}
\begin{tabularx}{\textwidth}{@{}p{2.7cm}X@{}}
\toprule
\textbf{Level} & \textbf{Expected capability} \\
\midrule
Minimal & Core envelope handling, participant lifecycle, task lifecycle, evidence log, error model, profile discovery, and basic schema validation. \\
Recommended & Minimal plus Review, Modes, stronger validation, production-grade storage, observability, operational runbooks, and tested recovery behaviour. \\
Full (planned) & Reserved for a future revision. Reaching Full requires implementation of every method in the catalogue, A2A composition, external evidence anchoring via \texttt{audit-scitt}, and successful execution of the published interop test suite against a second, independent implementation. The v0.2 reference implementations meet the method-coverage and interop requirements (both pass the published harness on the same wire); the remaining gates for a Full claim are exhaustive cross-implementation fixtures and the SCITT-anchoring conformance vectors, both of which the v0.2 cycle treats as work in progress. \\
\bottomrule
\end{tabularx}
\end{table}

Implementations already meeting the technical requirements of the planned Full level are welcome to publish a Recommended attestation with the additional implemented methods listed as an addendum; promotion to Full opens once the interop substrate is in place. Self-attestation must be honest about which methods are implemented versus declared; consumers should treat a method named in the catalogue but absent from the attestation as unavailable in that implementation.

\subsection{What to evaluate}

\chap{} implementations should be evaluated across protocol correctness, interoperability, operational reliability, security, and governance utility. Protocol correctness includes schema validation, method semantics, state transitions, idempotency, error handling, and evidence append behaviour. Interoperability includes multiple clients and servers exchanging envelopes without private assumptions. Operational reliability includes latency, throughput, storage durability, backpressure, replay, disaster recovery, and high availability.

Security evaluation includes signature verification, key rotation, scope enforcement, privileged operation checks, mode safety, and evidence-chain verification. Governance utility evaluates whether auditors, operators, reviewers, and model owners can answer practical questions from the evidence log without reconstructing events from screenshots, chat messages, or application-specific traces.

\subsection{Suggested evaluation questions}

A \chap{} deployment should be able to answer the following questions:

\begin{enumerate}[leftmargin=*]
    \item Who delegated this task, to whom, and under which workspace policy?
    \item Which agent version produced the artefact, and what tools, data sources, or peer agents did it cite?
    \item Was the artefact reviewed, approved, rejected, overridden, abstained from, or escalated?
    \item If overridden, what changed, who changed it, why, and under which policy reference?
    \item Did the task run in shadow, trial, or production mode?
    \item Did any participant abstain, and what category of abstention was declared?
    \item Was there a handoff, deliberation, route decision, pause, rollback, supersession, or mode transition?
    \item Can an authorised third party verify that the evidence chain was not silently altered?
\end{enumerate}

\section{Implementation Guide}
\label{sec:implementation}

This section is informative. It describes the v0.2 repository state and a practical adoption path for implementers. It does not replace the normative Core, profile, schema, or conformance definitions.

\subsection{Current v0.2 repository state}

The v0.2 repository (\url{https://github.com/BrightbeamAI/chap}) is best understood as a public Draft: suitable for review, experimentation, and prototype implementation, and increasingly suitable for early production pilots, but not yet a stable 1.0 standard. The repository includes draft schemas for envelopes, participants, tasks, workspaces, evidence, routing, and method catalogues; written specifications for the profile suite; two interoperable reference implementations (TypeScript and Python) covering Core and every shipped profile; a routing-aware playground with two human browser sessions and a local LLM; MCP and A2A transport adapters that expose every CHAP method to external clients; conformance scaffolding with canonical test vectors and a runnable harness; and worked end-to-end examples.

The method catalogue contains 39 method handlers across Core and the active profiles: seven normative Core methods plus two implementation-supplied bootstrap operations (\texttt{workspace.create}, \texttt{workspace.set\_profiles}), six in \texttt{review/1.0}, two in \texttt{whisper/1.0}, four in \texttt{deliberation/1.0}, three in \texttt{handoff/1.0}, seven in \texttt{control/1.0}, three in \texttt{routing/1.0}, two in \texttt{security-signed/1.0}, and three in \texttt{audit-scitt/1.0}. Both reference implementations implement all 39 method handlers and pass the same conformance harness against them. Implementations should advertise the supported method surface through \texttt{workspace.describe}, and self-attestations should distinguish implemented methods from declared ones (Section~\ref{sec:conformance}).

\subsection{Minimal implementation path}

A minimal implementation should support the following capabilities:

\begin{enumerate}[leftmargin=*]
    \item JSON-RPC-style request, response, and error handling over at least one transport.
    \item Workspace creation or static workspace configuration.
    \item Participant join, leave, and describe operations.
    \item Task creation, update, completion, and state projection.
    \item An append-only evidence log with ordered sequence numbers.
    \item Method validation and standard error responses.
    \item Profile discovery, even if only \texttt{core/1.0} is advertised.
\end{enumerate}

This is sufficient to build an internal human-agent task queue. From there, the recommended next profile is Review, followed by Modes for controlled agent rollout and Routing where assignment or review depth depends on risk, cost, confidence, deadline, or criticality.

\subsection{Recommended adoption path}

For most enterprise teams, adoption should proceed in stages:

\begin{description}[leftmargin=!,labelwidth=1.7cm]
    \item[Stage 1] Implement Core with a simple Coordinator, workspace descriptor, task lifecycle, and evidence log.
    \item[Stage 2] Add Review to capture approval, rejection, override, abstention, and escalation.
    \item[Stage 3] Add Modes to support safe rollout of new agents and agent versions.
    \item[Stage 4] Add Routing where cost, criticality, deadline, risk, confidence, or authority affects assignment and review depth.
    \item[Stage 5] Add Identity and Security-Signed when accountability crosses teams, vendors, or regulatory boundaries.
    \item[Stage 6] Add Audit-SCITT when offline, third-party-verifiable evidence is required.
\end{description}

\subsection{Storage choices}

A \chap{} implementation needs state storage and evidence storage. In small deployments, a relational database can hold workspace state, tasks, artefacts, and evidence entries. In higher-throughput deployments, an append-only log such as Kafka, Pulsar, or NATS JetStream can hold the evidence stream while a database holds current-state projections. Cold artefacts can be stored in object storage using content hashes, retention rules, and access-control policies.

The important property is not the storage product but the consistency boundary. A protocol-visible state transition and its evidence entry should be committed atomically or recoverably. Implementations should avoid states that cannot be explained from the evidence log.

\subsection{Transport choices}

\chap{} is transport-agnostic. WebSocket is appropriate for real-time collaborative user interfaces. HTTP with server-sent events is a pragmatic web-friendly option. HTTP polling can work for simple integrations. Message-broker bindings are appropriate for high-throughput systems, event-driven architectures, and internal platform integrations.

Regardless of transport, implementations should preserve envelope validation, idempotency, error handling, participant accountability, and evidence append semantics.

\subsection{Reference playground}

The repository includes a playground where two human browser sessions and a local LLM collaborate using the protocol. This is important because \chap{} is not only an abstract schema. It is intended to make collaboration events observable in real workflows. A useful playground should let implementers inspect envelopes, task transitions, review decisions, overrides, routing decisions, and evidence entries as they occur.

\section{Deployment Patterns}
\label{sec:deployment-patterns}

This section is informative. \chap{} does not mandate a deployment topology. It defines collaboration semantics that can be implemented in several architectures. The four common topologies below make different trade-offs around centralisation, audit consistency, latency, operational complexity, and cross-organisational trust.

\begin{figure}[!ht]
\centering
\includegraphics[width=0.85\textwidth]{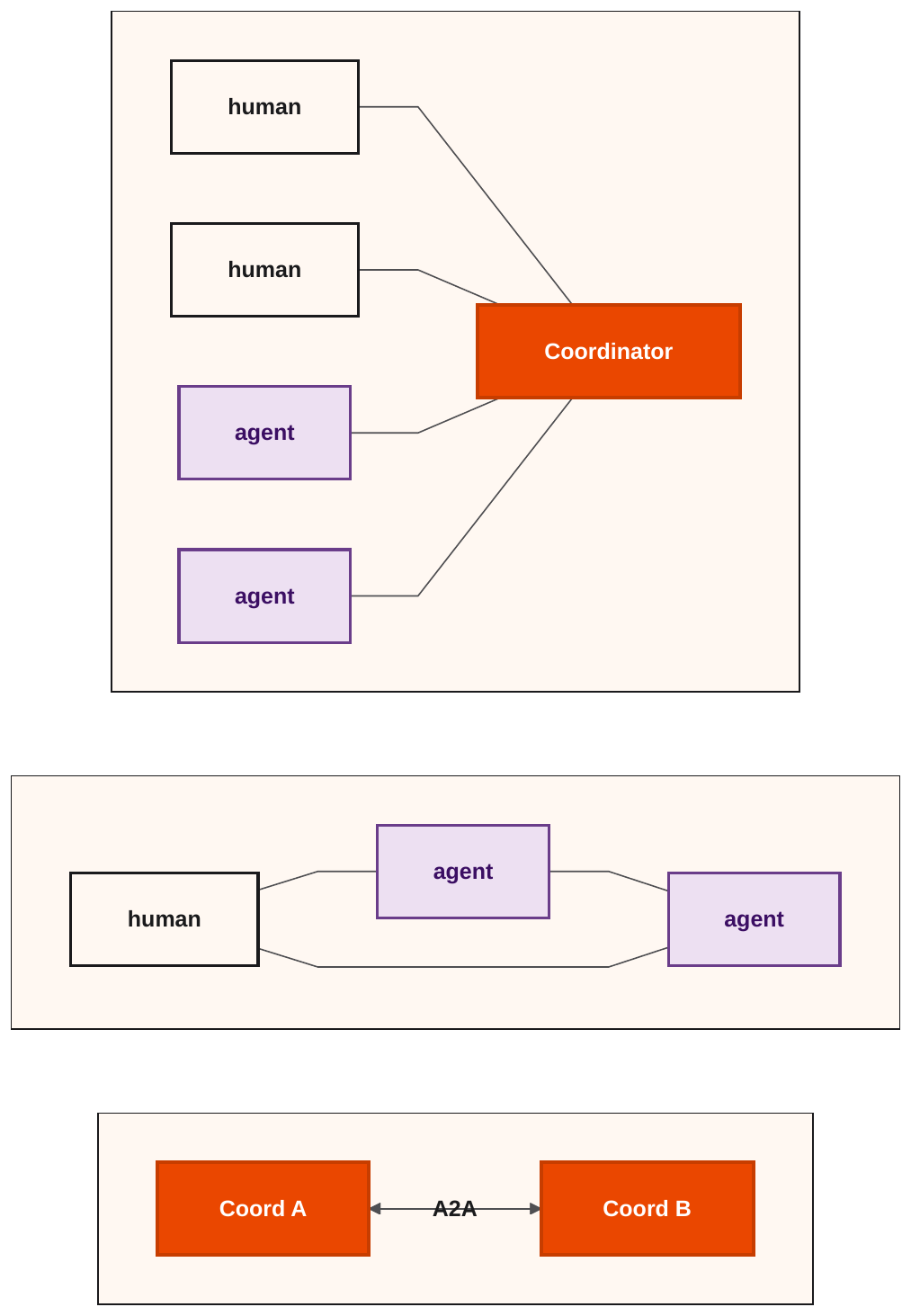}
\caption{Four deployment topologies. The Coordinator-mediated topology is the default. Peer-to-peer is possible but harder to secure and audit consistently. Federation links local workspaces through bridge participants and A2A. On-prem regulated deployments keep the Coordinator, evidence store, object store, and policy engine inside a controlled environment, with optional external SCITT anchoring for transparency.}
\label{fig:deployments}
\end{figure}

\subsection{Coordinator-mediated deployment}

The recommended default is a Coordinator-mediated deployment. All participants send envelopes to the Coordinator. The Coordinator validates, routes, applies workspace policy, and appends evidence. This topology simplifies policy enforcement, ordering, evidence consistency, monitoring, and operational support. It is also the easiest topology to secure and scale for early implementations.

\subsection{Peer-to-peer deployment}

A peer-to-peer topology is possible for advanced deployments, but it is harder to secure and audit consistently. Participants must agree on ordering, evidence append, conflict handling, policy enforcement, and trust assumptions. \chap{} can support such patterns, but they should not be the default for early or regulated implementations.

\subsection{Federated deployment}

A federated deployment uses local workspaces connected by bridge participants. Cross-organisational work can travel over A2A or another interoperation layer while local accountability remains in each \chap{} workspace. Federation is useful when each organisation needs to preserve its own policy boundary and audit trail. It also raises open questions around identity, policy translation, revocation, evidence sharing, confidentiality, and dispute resolution.

\subsection{On-prem regulated deployment}

A regulated deployment may place the Coordinator, evidence store, object store, and policy engine inside the organisation's controlled environment. Identity integrates with enterprise OIDC. Sensitive artefacts are stored externally with hashes and references in the \chap{} log. Signed envelopes and SCITT receipts can be enabled for non-repudiation and external verification. Operational dashboards can monitor override rates, abstentions, escalations, route decisions, mode transitions, policy violations, and incident signals.

\section{Relationship to Other Standards}
\label{sec:standards}

\chap{} follows a reuse-first design principle. It does not attempt to replace existing standards for transport, identity, signatures, canonicalisation, transparency, provisioning, tool access, or agent-to-agent interoperability. Instead, it defines the collaboration vocabulary that is missing between those layers.

For message exchange, \chap{} uses a JSON-RPC-2.0-inspired~\cite{jsonrpc} request, response, notification, and error pattern. For deterministic bytes when computing hashes or signatures, it uses JSON Canonicalization Scheme (JCS, RFC~8785)~\cite{jcs}. For structured override diffs, it uses JSON Patch (RFC~6902)~\cite{jsonpatch}. For enterprise human identity, it can integrate with OpenID Connect~\cite{oidc}, OAuth~2.0~\cite{oauth}, DPoP~\cite{dpop}, and token-to-key binding via proof-of-possession (RFC~7800)~\cite{rfc7800}. For richer cross-organisational claims, it can bind participants to W3C Verifiable Credentials~\cite{vc20}. For transparency-backed audit, it profiles SCITT~\cite{scitt} over signed statements (using COSE~\cite{cose} as the underlying signature container). For service identity, it can integrate with SPIFFE/SPIRE~\cite{spiffe}. For user provisioning, it can integrate with SCIM~\cite{scim}. For agent-to-tool calls inside a workspace, it composes with MCP~\cite{mcp}. For agent-to-agent delegation across organisational boundaries, it composes with A2A~\cite{a2a}. The composition with MCP and A2A runs in both directions: a Coordinator can cite MCP tool calls and A2A exchanges outward, and can also present itself as an MCP server or A2A agent inward, with every CHAP method exposed as a tool or skill. Section~\ref{sec:integration-packages} describes the inward direction in more detail.

The contribution of \chap{} is the workspace collaboration layer: participants, tasks, artefacts, evidence entries, task lifecycle events, review decisions, structured overrides, abstentions, escalations, handoffs, deliberations, mode transitions, and profile negotiation. These are the recurring human-agent work primitives that existing adjacent standards do not define as a shared protocol surface.

\section{Informative User Journeys}
\label{sec:user-journeys}

This section turns the protocol model into concrete journeys. The journeys are informative rather than normative. They illustrate the collaboration events that \chap{} makes explicit and show how Core and profiles can be combined in realistic deployments.

\subsection{Journey 1: minimal Core flow}

The smallest useful \chap{} flow has two participants and one task. Alice joins a workspace. A triage bot joins. Alice describes the workspace, creates a task, the bot records progress, completes the task, and Alice reads the evidence log. In the v0.2 reference implementation, this flow uses \texttt{participant.join}, \texttt{workspace.describe}, \texttt{task.create}, \texttt{task.update}, \texttt{task.complete}, and \texttt{audit.read}.

\begin{lstlisting}[caption={Minimal Core task creation in the reference style.},label={lst:taskcreate}]
{
  "jsonrpc": "2.0",
  "id": "4",
  "method": "task.create",
  "params": {
    "workspace": "wsp_demo",
    "from": "human:alice@example.org",
    "to": "agent:triage-bot",
    "ts": "2026-05-17T09:01:00Z",
    "kind": "draft_response",
    "assignee": "agent:triage-bot",
    "input": {
      "ticket_id": "INC-48219",
      "customer_message": "Where is my order?"
    }
  }
}
\end{lstlisting}

The evidence log becomes the protocol record for the workspace. Even this minimal deployment is useful for internal team bots, solo agent workbenches, and simple structured task queues.

\subsection{Journey 2: drafter-reviewer}

A document drafting agent generates a response, but the workspace policy requires human approval before delivery. The Coordinator opens a review request. The reviewer can approve the draft, reject it with a reason, override it with a structured diff, abstain, or escalate. \chap{} preserves the distinction between the agent's draft, the human's final version, and the rationale for the change.

\subsection{Journey 3: override as learning data}

A claims-processing agent drafts a coverage explanation and assigns it a \texttt{logical\_id} for the claim. A human reviewer edits the explanation to add missing policy nuance, tags the override as \texttt{policy-exception} and \texttt{customer-specific-context}, carries the same \texttt{logical\_id} forward, and sets \texttt{intent\_preserved} to \texttt{true}; the human kept the underlying coverage decision but refined how it was explained. Over time, the organisation can query overrides by tag, model version, task kind, reviewer role, risk tier, and policy reference. The \texttt{intent\_preserved} signal separates two failure modes: a high refining-override rate around one policy clause may indicate the agent reaches the right decision in the wrong way (weak retrieval or unclear template), while a high substituting-override rate on the same clause indicates the agent reaches the wrong decision (ambiguous policy or insufficient task context). These tune to different fixes.

\subsection{Journey 4: abstain and escalate}

An agent encounters low confidence and conflicting evidence. Instead of generating a risky answer, it calls \texttt{abstain.declare} with a category such as \texttt{insufficient\_evidence}, \texttt{policy\_conflict}, or \texttt{authority\_boundary}. The Coordinator records the abstention and opens an escalation to the appropriate human or group. Abstention rate then becomes an operational signal: too many abstentions may reveal missing tools, incomplete knowledge bases, unclear policies, or poorly calibrated autonomy limits.

\subsection{Journey 5: whisper during task execution}

A support agent is midway through a task and finds an ambiguity: a customer has two active orders but mentions cancellation of only one. The agent asks a \texttt{whisper.ask} question with typed options and a default if the human does not answer within the configured time window. The answer is recorded, and the agent completes the task. This avoids both unnecessary full review and unsafe silent assumption.

\subsection{Journey 6: shift handoff}

At the end of a shift, a human operator proposes a handoff to another participant. The handoff includes task summaries, current status, next action, blockers, and active policy constraints. The receiving participant accepts, and the Coordinator reassigns the relevant tasks. Unlike informal chat notes, the handoff becomes part of the workspace evidence chain.

\subsection{Journey 7: multi-human deliberation}

A proposed goodwill credit exceeds an individual reviewer's authority. The workspace opens a deliberation with a \texttt{weighted\_vote:2.0} rule. Authorised reviewers vote, the deliberation closes when the threshold is met, and the result becomes an artefact referenced by the downstream credit-issuance task. This makes organisational accountability explicit without hard-coding one decision structure into the protocol.

\begin{figure}[!ht]
\centering
\includegraphics[width=\textwidth]{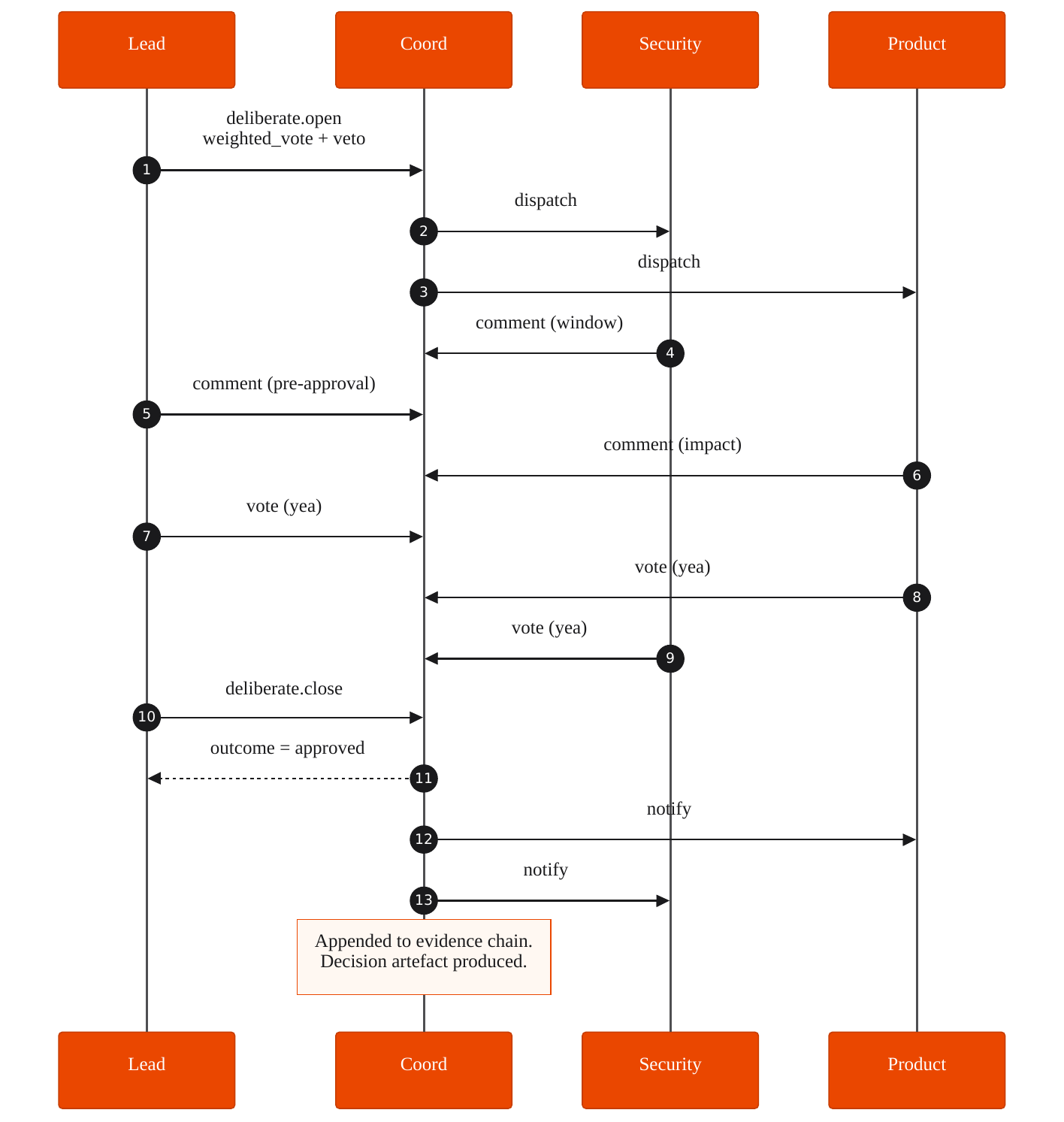}
\caption{Weighted-vote deliberation. The Coordinator opens deliberation when the proposed credit exceeds the individual-reviewer authority threshold. Authorised reviewers cast weighted votes. The Coordinator closes the deliberation when the configured rule is satisfied. The outcome becomes an artefact that can be cited by the downstream credit-issuance task.}
\label{fig:deliberation}
\end{figure}

\subsection{Journey 8: MCP tool-using agent}

An agent calls three MCP tools: order lookup, customer history, and knowledge-base retrieval. \chap{} does not duplicate the MCP transcript. Instead, the agent's artefact cites each relevant tool call with input and output hashes. The workspace evidence chain can therefore show which external evidence influenced the artefact without absorbing all tool traffic into \chap{}.

\subsection{Journey 9: A2A bridge participant}

A local workspace delegates specialised analysis to an agent owned by another organisation. The remote peer communicates over A2A. Locally, a \texttt{service:bridge} participant represents the remote work inside the \chap{} workspace. The bridge signs local \chap{} envelopes, links them to A2A task identifiers, and returns a completion artefact when the remote peer responds. The local audit remains coherent while the cross-organisational agent traffic remains in A2A.

\subsection{Journey 10: driving a workspace from an MCP client or A2A orchestrator}

A solution architect wants to demonstrate a CHAP workflow without writing application code. They configure Claude Desktop to load the CHAP MCP reference server. The server wraps a Coordinator and exposes every CHAP method as a tool named \texttt{chap.<method>}. The architect types: ``create a workspace called \texttt{wsp\_demo}, join me as a human owner and a drafting agent, then create a draft task for the agent.'' Claude Desktop calls \texttt{chap.workspace.create}, \texttt{chap.participant.join} twice, and \texttt{chap.task.create}. The architect inspects the resulting audit chain by asking for it; Claude Desktop calls \texttt{chap.audit.read}. The same audit chain that a coded client would have produced lands in the workspace, with every envelope hash-linked.

The same pattern works for an A2A orchestrator. The orchestrator discovers the Coordinator's Agent Card at the well-known URL, sees 39 skills named \texttt{chap.<method>}, and delegates work to them. The audit chain is identical to the chain a direct JSON-RPC client would have produced. This is the inward composition direction described in Section~\ref{sec:integration-packages}: the protocol's collaboration vocabulary becomes available to any caller fluent in MCP or A2A.

\section{Limitations and Future Work}
\label{sec:limitations}

\chap{} v0.2 is a public Draft. It is suitable for review, prototyping, early production pilots, and implementation feedback, but several issues must mature before a stable 1.0 release.

\begin{enumerate}[leftmargin=*]
    \item \textbf{Cross-implementation interop test coverage.} The TypeScript and Python reference implementations now pass the same conformance harness against the published test vectors. The next step is to broaden the published interop test suite with profile-specific fixtures and negative tests, and to add a third reference implementation built by an independent team to harden the conformance ladder.

    \item \textbf{Confidentiality of evidence content.} The current design supports hashes and opaque artefact references, but a full confidentiality profile for encrypted evidence, selective disclosure, controlled sharing, and redaction remains future work. The interaction with deletion regimes such as GDPR Article 17 (right to erasure) on an append-only chain is part of this work.

    \item \textbf{Formal interoperability test suite.} The repository includes canonical test vectors for signing, canonicalisation, and evidence chaining plus a runnable harness, but cross-implementation fixtures, negative tests, and profile-specific test cases must mature before 1.0 stability.

    \item \textbf{Semantic policy portability.} \chap{} can record policy identifiers, routing decisions, and review outcomes, but it does not standardise policy languages, risk thresholds, or decision criteria. The intentional non-goal (Section~\ref{sec:non-goals}) preserves the protocol's neutrality; the practical consequence is that two \chap{} deployments may carry equivalent evidence without being semantically interoperable.

    \item \textbf{Human factors.} Structured review and override capture must be designed so that they improve accountability without overburdening reviewers, slowing operations, or encouraging mechanical approval.

    \item \textbf{Override quality.} Human overrides are valuable signals, but they are not infallible labels. The \texttt{intent\_preserved} flag added in v0.2 helps separate refinement from substitution, but reviewer consistency, local bias, calibration, disagreement, and the conditions under which overrides can safely inform model or workflow improvement remain open research questions.

    \item \textbf{Federation depth.} Cross-workspace and cross-organisation collaboration need deeper treatment of identity, policy translation, revocation, evidence exchange, confidentiality, and dispute resolution. Hierarchical-workspace composition (intra-organisational parent/child semantics distinct from cross-organisational federation) is an explicit candidate for v0.3.

    \item \textbf{Long-lived cryptographic evidence.} Ed25519 is practical for current deployments, but long-lived evidence chains may require additional signature suites, key-rotation practices, archival verification, and hybrid post-quantum options.

    \item \textbf{Empirical evaluation.} \chap{} needs multi-implementation pilots and comparative studies across sectors, task types, risk levels, and deployment topologies.

    \item \textbf{Profile churn.} Profiles are expected to evolve faster than Core. Deployments pinning specific profile versions (\texttt{review/1.0}, \texttt{routing/1.0}) will gain stability; deployments wanting the latest semantics will need a profile-upgrade discipline that the spec does not currently formalise.
\end{enumerate}

\section{Conclusion}
\label{sec:conclusion}

The next phase of useful AI will not be defined by model capability alone. It will depend on whether humans, agents, and services can work together in ways that are explicit, governable, and replayable. Without a protocol for this collaboration layer, organisations will continue to rebuild the same primitives (task assignment, review, override, abstention, escalation, handoff, deliberation, and audit) inside isolated applications. That fragmentation limits interoperability, weakens governance, and makes it harder to learn from the human decisions that keep agentic systems aligned with operational reality.

\chap{} proposes a compact and composable protocol for the shared human-agent workspace. Core is intentionally small: workspaces, participants, tasks, artefacts, evidence entries, and a message envelope. Profiles add the capabilities that production deployments need: review, modes, routing, whisper prompts, handoff, deliberation, control, identity, signatures, and transparency-backed audit. The design reuses existing standards wherever possible and adds new vocabulary only where the missing verbs of human-agent collaboration require it.

This v0.2 draft is not a final standard. It is a concrete starting point: draft schemas, profile specifications, two interoperable reference implementations in TypeScript and Python, MCP and A2A transport adapters, examples, a playground, and conformance scaffolding. The next step is a third independent implementation, broader cross-implementation interop testing, security review, and empirical deployment in real human-agent workflows. The Human--Agent Symphony will not emerge simply because models improve. It will emerge when the collaboration layer between humans and agents becomes explicit, observable, portable, and shared. \chap{} is a proposal for that layer.

\appendix

\section{Example End-to-End Audit Trace}
\label{app:audit-trace}

This appendix is informative. It illustrates how a support workflow involving refund review, override, deliberation, and tool-using credit issuance could appear when replayed from the evidence log.

\begin{lstlisting}[caption={Illustrative \chap{} audit replay.},label={lst:audittrace}]
seq=2012  task.assign         human:alice       -> agent:triage-bot       INC-48910
seq=2013  task.accept         agent:triage-bot  -> service:coordinator
seq=2014  task.start          agent:triage-bot  -> service:coordinator
seq=2015  whisper.ask         agent:triage-bot  -> human:alice            refund vs replacement?
seq=2016  whisper.answer      human:alice       -> agent:triage-bot       refund_per_request
seq=2017  task.complete       agent:triage-bot  -> service:coordinator    draft + MCP citations
seq=2018  review.request      service:coordinator -> human:alice          review required
seq=2019  decide.override     human:alice       -> service:coordinator    tone + credit change
seq=2020  deliberate.open     service:coordinator -> group:credit-board    authorise credit?
seq=2021  deliberate.vote     human:bob         -> service:coordinator    yea
seq=2022  deliberate.vote     human:carol       -> service:coordinator    yea
seq=2023  deliberate.close    service:coordinator -> workspace            approved
seq=2024  task.assign         service:coordinator -> agent:credit-issuer   issue credit
seq=2025  task.complete       agent:credit-issuer -> service:coordinator   credit transaction id
seq=2026  notify.message      service:coordinator -> human:alice           final response sent
\end{lstlisting}

\section{Starter Profile Bundles}
\label{app:profile-bundles}

Table~\ref{tab:bundles} gives informative starter bundles for common deployment patterns.

\begin{table}[h]
\centering
\caption{Suggested starter profile bundles.}
\label{tab:bundles}
\begin{tabularx}{\textwidth}{@{}p{3.3cm}X@{}}
\toprule
\textbf{Use case} & \textbf{Suggested profiles} \\
\midrule
Internal team chatbot & Core only, optionally Review. \\
Document drafting with approval & Core + Review + Modes. \\
Support triage & Core + Review + Whisper + Handoff + Routing. \\
Regulated approval workflow & Core + Review + Modes + Identity-OIDC or Identity-VC + Security-Signed + Audit-SCITT. \\
Cross-organisation specialist delegation & Core + Review + Identity-VC + Security-Signed + A2A bridge pattern + Audit-SCITT. \\
Incident response & Core + Handoff + Deliberation + Control + Security-Signed. \\
\bottomrule
\end{tabularx}
\end{table}

\section{Implementation Checklist}
\label{app:implementation-checklist}

This checklist summarises practical implementation checks for early \chap{} deployments.

\begin{enumerate}[leftmargin=*]
    \item Validate envelope shape and method parameters against schemas.
    \item Enforce participant membership, scopes, and roles.
    \item Implement idempotency for repeated message identifiers.
    \item Append evidence atomically with accepted state transitions.
    \item Preserve rejected-envelope errors for operational debugging where appropriate.
    \item Expose \texttt{workspace.describe} and profile discovery.
    \item Keep task projection derivable from the evidence log.
    \item Store artefacts by content hash or immutable URI where possible.
    \item Separate routing hints from routing policy.
    \item Capture override diffs, rationale, tags, and policy references.
    \item Define mode promotion and demotion procedures before production use.
    \item Monitor override, rejection, abstention, escalation, and incident rates.
    \item Add identity and signing before crossing organisational or regulatory trust boundaries.
    \item Add external transparency receipts when third-party verification is required.
\end{enumerate}

\section{CHAP in practice}
\label{app:applications}

This appendix walks through twelve situations where teams reach for the protocol. They run from one person on a side project up to GMP-regulated manufacturing. They are not meant to be exhaustive. They are meant to be recognisable. If one describes a reader's week, the rest of the report makes more sense.

A note on scope before the cases. \chap{} defines what gets recorded and how it links together. It does not pick the model, write the prompts, design the routing rules, interpret the regulator, or decide whether a particular human review was substantively good enough. Each case below describes what \chap{} contributes; the substantive work above it remains the deploying organisation's.

The code samples below use the TypeScript \texttt{@chap/coordinator} package. The Python \texttt{chap-coordinator} package exposes the same surface and the same envelope shapes; implementations in other languages follow the same wire format. The content-type for HTTP transports is \texttt{application/chap+json} and any HTTP client works.

\subsection{The solo developer who can't remember what they overrode}

A single developer ships to GitHub. Cursor reviews every pull request. The developer accepts most of its suggestions, rejects some, rewrites a few. Three months in they have a vague sense that the bot is broadly useful but couldn't say with any precision which classes of suggestion it gets wrong.

A coordinator running locally with embedded storage is twenty lines of setup:

\begin{lstlisting}[caption={Bootstrap a local coordinator.},label={lst:case1-bootstrap}]
import { Coordinator } from "@chap/coordinator";

const coord = new Coordinator({ storage: "sqlite:./chap.db" });
coord.dispatch({
  jsonrpc: "2.0", id: "init",
  method: "workspace.create",
  params: { workspace_id: "wsp_my_reviews",
            profiles: ["core/1.0", "review/1.0"] }
});
coord.dispatch({
  jsonrpc: "2.0", id: "j1",
  method: "participant.join",
  params: { workspace_id: "wsp_my_reviews",
            uri: "human:me@local", type: "human" }
});
\end{lstlisting}

Wired into the PR script, every Cursor review becomes an artefact on a task. A rejection is a \texttt{decide.reject} with a category. An edit before merging is a \texttt{decide.override} with the diff and a one-line rationale.

\begin{lstlisting}[caption={Capturing an override.},label={lst:case1-override}]
{
  "method": "decide.override",
  "params": {
    "task_id": "tsk_pr_482",
    "from": "human:me@local",
    "logical_id": "lgl_pr_482_review",
    "intent_preserved": true,
    "diff": [
      { "op": "replace", "path": "/comments/0/severity",
        "from": "warning", "to": "info" }
    ],
    "rationale": "Bot flags unused parameter on every event handler. False positive in this codebase: handlers conform to a framework signature.",
    "tags": ["false-positive", "framework-pattern-misread"]
  }
}
\end{lstlisting}

Two months in, the override analyser surfaces a tag histogram showing 31 false-positive overrides, 22 of which trace to the same framework pattern. The next prompt revision for Cursor cites the pattern by name.

\subsection{Marketing copy with one drafter and one editor}

A two-person marketing function at an early-stage company composes the workflow: one writes, one edits, one approves; an agent produces a first draft from the brief; the drafter refines, the editor signs off. The agent is competent. Some weeks it is better than competent. But the editor keeps making the same kinds of edits, softening corporate openers, swapping passive constructions, killing one cliché in particular, and the team retunes the agent's prompt every Friday based on what they remember of the week.

With Core and \texttt{review/1.0}, each brief is a task. The agent's draft is an artefact. The editor's revisions become overrides with tags chosen from a short list.

\begin{lstlisting}[caption={Editor override with tag taxonomy.},label={lst:case2-override}]
{
  "method": "decide.override",
  "params": {
    "task_id": "tsk_brief_acme_q3_2026",
    "from": "human:editor@studio.com",
    "logical_id": "lgl_brief_acme_q3_2026",
    "intent_preserved": true,
    "diff": [
      { "op": "replace", "path": "/sections/0/text",
        "from": "Industry-leading solutions for forward-thinking teams.",
        "to": "We help teams ship faster. Here's how." }
    ],
    "rationale": "Opener was generic corporate boilerplate.",
    "tags": ["opener-rewritten", "tone-corporate-to-warm"]
  }
}
\end{lstlisting}

After two months, the tag distribution shows \texttt{opener-rewritten} at 62 per cent. The next prompt revision bans three specific opener patterns by name; the cliché count drops the following month, and the team sees the drop in the data rather than remembering it from feel. The \texttt{modes/1.0} profile is useful here too: the agent stays in \emph{trial} until the tag distribution looks stable, then moves to \emph{production} where the editor spot-checks rather than gates.

\subsection{The indie founder, the inbox, and the angry customer}

A founder runs a small SaaS product alone. Support volume has crossed the threshold where they cannot read every ticket, so an AI triage agent now drafts responses; the founder approves, edits, or escalates. The pipeline is a webhook, the OpenAI API, and Linear.

Six weeks in, a customer files a chargeback citing a refund policy the bot got wrong. The customer's original email is recoverable. The final reply is recoverable. The bot's intermediate draft is in OpenAI logs somewhere, but only for thirty days, and keys have been rotated since. The founder's reasoning for accepting the draft is in nobody's head except their own.

The remedy is to run every ticket through a coordinator and replay the chain when contested:

\begin{lstlisting}[caption={Audit-read query for a disputed ticket.},label={lst:case3-replay}]
const audit = await coord.dispatch({
  jsonrpc: "2.0", id: "a1",
  method: "audit.read",
  params: {
    workspace_id: "wsp_support",
    filter: { task_id: "tsk_ticket_chargeback_8821" }
  }
});

for (const entry of audit.result.entries) {
  console.log(`[${entry.envelope.ts}] ${entry.envelope.method}`);
  console.log(`  from: ${entry.envelope.params.from}`);
  console.log(`  ${JSON.stringify(entry.envelope.params).slice(0, 120)}`);
}
\end{lstlisting}

The output is the whole story in chronological order. Looking back, the founder also notices that seven other tickets had the same wrong policy quoted by the same bot version. The agent's retrieval is fixed to consult the actual policy document; the eighth ticket is correct. Six months later, when the founder hires a support contractor, the workspace policy is in \texttt{workspace.describe} and the override patterns are visible in the audit log. Onboarding is reading the chain rather than reading the founder's mind.

\subsection{Support operations at 03:00}

A mid-size SaaS company operates a support function of around forty agents across three time zones. Tier-1 humans clear the queue with AI-drafted responses. Tier-2 specialists handle escalations. A senior reviewer gate-keeps refunds above a monetary threshold. The team uses Zendesk for tickets, Slack for context, and an in-house drafting agent.

It is 03:00 in EMEA. A shift lead is taking over from the Asia-Pacific team. They open Slack and find a 200-message thread with overnight context, several updates that scrolled past midnight, and a note pinned at the top referring to Linear for blockers. Linear's blocker tickets reference Zendesk tickets that have to be looked up. Twenty minutes pass before they have the picture.

With \chap{}, the Asia-Pacific shift lead's last act is one envelope:

\begin{lstlisting}[caption={Shift handoff envelope.},label={lst:case4-handoff}]
{
  "method": "handoff.propose",
  "params": {
    "from": "human:apac.lead@acme.com",
    "to": ["group:emea-shift"],
    "tasks": ["tsk_ticket_8821", "tsk_ticket_8847", "tsk_ticket_8903"],
    "note": "8821: awaiting customer reply, no action needed. 8847: escalated to legal review around 23:00, they're on it. 8903: agent v2.4 drafted but quoted yesterday's shipping policy, do not approve as-is. We logged the wrong-policy-quoted tag three times in the last shift, looks like the agent hasn't picked up the policy refresh. Pinged eng."
  }
}
\end{lstlisting}

The EMEA lead picks up the queue with a single query, then reads the handoff note alongside each task:

\begin{lstlisting}[caption={Picking up the queue at shift change.},label={lst:case4-pickup}]
const assignments = await coord.dispatch({
  jsonrpc: "2.0", id: "q1",
  method: "audit.read",
  params: {
    workspace_id: "wsp_support",
    filter: {
      method: "handoff.accept",
      from: "group:emea-shift",
      since: shiftStart
    }
  }
});

for (const a of assignments.result.entries) {
  const taskId = a.envelope.params.task_id;
  const task = await coord.dispatch({
    jsonrpc: "2.0", id: nextId(),
    method: "task.describe",
    params: { workspace_id: "wsp_support", task_id: taskId }
  });
  console.log(`${task.result.id}: ${task.result.handoff_note}`);
}
\end{lstlisting}

The note is permanent. It does not scroll, does not expire, and is in the audit log against those task identifiers. The wider operation composes accordingly: \texttt{routing/1.0} decides which tickets need senior eyes based on refund size and risk tier; \texttt{deliberation/1.0} routes refunds above a higher threshold to a two-reviewer vote; \texttt{whisper/1.0} lets the agent ask the Tier-1 handler a typed question (``refund or replacement?'') mid-draft. Most teams already have routing, deliberation, and clarifying questions; the difference is having them in one chain rather than across four tools and one chat channel.

\subsection{Sixty engineers and a code-review bot they don't trust the same way}

A Series-B-stage startup of around sixty engineers in four squads runs AI code review on every pull request (Greptile, Cursor's review feature, or an in-house equivalent). Adoption is uneven. One squad trusts the bot deeply. Another has muted it. The security squad asks, one Wednesday: \emph{did the bot ever flag a real security issue that the team dismissed?} There is no answer; dismissals leave no trace.

\chap{} turns dismissals into first-class events.

\begin{lstlisting}[caption={Dismissal with category and rationale.},label={lst:case5-reject}]
{
  "method": "decide.reject",
  "params": {
    "task_id": "tsk_pr_review_12482",
    "from": "human:alice@acme.com",
    "reason_category": "false-positive",
    "rationale": "Flagged SQL injection on line 47. Query is parameter-bound through sqlx; bot misread the template-string interpolation.",
    "based_on_artefact_id": "art_bot_review_..."
  }
}
\end{lstlisting}

The security squad's quarterly audit query is a few lines:

\begin{lstlisting}[caption={Quarterly dismissal-rate query.},label={lst:case5-query}]
const dismissals = await coord.dispatch({
  jsonrpc: "2.0", id: "sec-q1",
  method: "audit.read",
  params: {
    workspace_id: "wsp_eng_reviews",
    filter: {
      method: "decide.reject",
      since: "2026-01-01T00:00:00Z",
      until: "2026-04-01T00:00:00Z",
      artefact_tags: ["security-flag"]
    }
  }
});

const byCategory = {};
for (const e of dismissals.result.entries) {
  const cat = e.envelope.params.reason_category;
  byCategory[cat] = (byCategory[cat] || 0) + 1;
}
\end{lstlisting}

Two profiles carry most of the weight. The \texttt{audit-scitt/1.0} profile anchors the workspace evidence to a transparency log, which lets the security team verify dismissal history without holding write access to engineering's tools. The \texttt{modes/1.0} profile lets the security squad pin \emph{production} mode on security-sensitive paths (every flagged comment must be acted on or dismissed with a reason) while everything else stays in \emph{trial}. A quarter later, the security team has evidence for its own internal audit; the bot's prompts get tuned for the false-positive patterns; the muted squad starts re-enabling the bot.

\subsection{A junior, a partner, and a contract that ships tonight}

An in-house legal team, or a small firm, has a junior associate running first-pass review on a master services agreement with an AI assistant. For non-standard terms, the file moves to a partner.

It is 18:00 on a Tuesday. The agreement needs to ship by 21:00. The junior approves what looks like a standard liability cap. The partner, reviewing at 19:30, catches it: Acme Corp is a Tier-A client, and Tier-A clients have a non-negotiable cap structure the junior did not recognise. The partner rewrites the clause.

\begin{lstlisting}[caption={Override that substitutes a different decision.},label={lst:case6-override}]
{
  "method": "decide.override",
  "params": {
    "task_id": "tsk_msa_acme_corp",
    "from": "human:smith.j@firm.com",
    "logical_id": "lgl_msa_acme_corp",
    "intent_preserved": false,
    "diff": [
      { "op": "replace", "path": "/clauses/liability/cap",
        "from": "12_months_fees", "to": "24_months_fees" },
      { "op": "add", "path": "/clauses/liability/carve_outs",
        "value": "IP-indemnity-uncapped" }
    ],
    "rationale": "Acme is Tier-A; the 12-month cap shouldn't have left juniors' desks. Updating to the Tier-A structure. This is an escalation miss, junior should have flagged when they saw the client name.",
    "tags": ["liability-cap-modified", "tier-a-client", "junior-escalation-miss"],
    "policy_refs": ["firm.contract-policy.v3#tier-a-clients"]
  }
}
\end{lstlisting}

\texttt{intent\_preserved: false} is the load-bearing field. It tells any downstream analytics that this override substituted a different decision rather than refining the junior's expression of the same one. Across a quarter, the \texttt{junior-escalation-miss} tag count is the firm's most actionable training signal. The \texttt{identity-vc/1.0} profile is doing related work: the partner's authority to approve a Tier-A modification is a credential issued by bodies outside the workspace (bar admission, partnership status), and the signing operation carries the credential by reference.

\subsection{Trust and Safety, three auditors, one queue}

A community platform serving roughly ten million users runs the usual queue: reported content arrives, an AI classifier triages it (hate, harassment, child-safety violations, violence), human reviewers adjudicate.

The platform answers to three audit demands at once. The EU Digital Services Act and the UK Online Safety Act both require statements of reasons for takedowns and meaningful human review on harder calls. Law-enforcement subpoenas ask for specific user histories. Creators appeal takedowns through an internal process that needs the original decision record. Today these demands are served by three logging pipelines that don't quite agree with each other.

A \chap{} workspace replaces those three pipelines with one chain.

\begin{lstlisting}[caption={Takedown decision with policy citations.},label={lst:case7-decide}]
{
  "method": "decide.approve",
  "params": {
    "task_id": "tsk_report_19204821",
    "from": "human:reviewer.j@platform.com",
    "based_on_artefact_id": "art_ai_classification_...",
    "decision": "take-down",
    "policy_refs": ["community.guidelines.v18#harassment", "DSA.art.16"],
    "rationale": "Targeted harassment of named individual; pattern matches s.3.2; three other accounts reported same user this week.",
    "tags": ["harassment", "targeted", "pattern-of-behaviour"]
  }
}
\end{lstlisting}

The DSA statement of reasons becomes a templating exercise against one source of truth, not a reconciliation across three pipelines:

\begin{lstlisting}[caption={Generating a DSA statement of reasons.},label={lst:case7-statement}]
async function generateDSAStatement(taskId: string): Promise<string> {
  const chain = await coord.dispatch({
    jsonrpc: "2.0", id: nextId(),
    method: "audit.read",
    params: { workspace_id: "wsp_ts", filter: { task_id: taskId } }
  });
  const decision = chain.result.entries.find(
    e => e.envelope.method === "decide.approve"
  ).envelope.params;
  const classification = await getArtefact(decision.based_on_artefact_id);
  return `
Statement of Reasons (DSA Article 17)
Content removed under: ${decision.policy_refs.join(", ")}
Decision date: ${decision.ts}
Classification: ${classification.category}
Human review: ${decision.from}
Reasoning: ${decision.rationale}
`;
}
\end{lstlisting}

The subpoena response is the relevant slice of the chain. The appeal references the original decision as its base. \texttt{audit-scitt/1.0} is load-bearing because regulators want third-party-verifiable evidence and law-enforcement subpoenas need verifiable chain integrity. \texttt{deliberation/1.0} carries deplatforming decisions on users with material public reach. \texttt{modes/1.0} rolls out new classifier versions safely.

\subsection{The creative shop and the client who ``never approved that''}

A fifteen-person creative agency uses AI for first drafts of copy, images, and storyboards. The work goes through three internal eyes before the client sees it; the client approves; the work ships.

Six months later the campaign is running and the client's new chief marketing officer is furious. The tagline is not what they signed off on, they say. The agency has a chat thread from May with the original brief, design-tool comments from June with the internal revision rounds, an email from July with the version that went to the client, and a reply two days later that says ``looks great let's go''. Nobody can find what exactly the new CMO is referring to, and the approval email was from the CMO's predecessor, who left the company in September.

The relevant \chap{} move is \texttt{security-signed/1.0} composed with \texttt{identity-oidc/1.0}. The client approval is signed with an OIDC-bound key, references the final artefact by content hash, and is non-repudiable.

\begin{lstlisting}[caption={Client-side signing flow.},label={lst:case8-sign}]
import { signEnvelope } from "@chap/client";

const envelope = {
  jsonrpc: "2.0", id: nextId(),
  method: "decide.approve",
  params: {
    workspace_id: "wsp_agency_acme_q4",
    task_id: "tsk_campaign_q4_launch",
    from: clientOidcSubject,
    based_on_artefact_id: "art_final_v7_...",
    logical_id: "lgl_campaign_q4_launch_final",
    content_hash: "sha256:7f8e9d0c..."
  }
};

const signed = await signEnvelope(envelope, clientOidcKey);
await postEnvelope(coordinatorUrl, signed);
\end{lstlisting}

The chain shows precisely which version was approved, by which named individual, at which timestamp, against which brief. The content hash means any change to the artefact bytes invalidates the signature on the approval; the dispute resolves itself. The internal-review overrides with tags such as \texttt{tone}, \texttt{imagery}, \texttt{factual-correction}, \texttt{client-preference} build a portrait over time of which AI tools produce work in which classes of brief.

\subsection{The internal Q\&A bot that keeps getting the same thing wrong}

A five-hundred-person organisation operates an internal retrieval-augmented question-answering bot covering benefits, IT setup, expense policy, and internal tooling. It answers roughly seventy per cent of questions adequately. The remainder escalate to subject-matter experts.

The SME loop is broken in a specific way. An employee asks about parental-leave eligibility for adoption. The bot does not know. The question is routed to HR. The HR partner answers in email; the employee is satisfied; the conversation ends. Three weeks later a different employee asks the same question.

The pattern that helps is \texttt{whisper/1.0}. When the bot is uncertain, it whispers a typed question to the SME team:

\begin{lstlisting}[caption={Whisper with typed options and a default.},label={lst:case9-whisper}]
{
  "method": "whisper.ask",
  "params": {
    "task_id": "tsk_employee_q_8821",
    "from": "agent:hr-bot",
    "to": ["group:hr-team"],
    "question": "Employee is asking about parental-leave eligibility for adoption. Policy doc doesn't explicitly say. Do we treat adoption same as biological?",
    "options": ["yes-same", "no-different", "needs-case-by-case"],
    "default_if_no_answer": "needs-case-by-case",
    "deadline": "2026-05-17T17:00:00Z"
  }
}
\end{lstlisting}

The HR partner taps \texttt{yes-same} in thirty seconds. Inside the bot's run loop, the answer is captured back into the knowledge base so the question is not asked again:

\begin{lstlisting}[caption={Bot consumes whisper response into its policy KB.},label={lst:case9-loop}]
const whisperResp = await coord.dispatch({
  jsonrpc: "2.0", id: nextId(),
  method: "whisper.ask",
  params: { /* as above */ }
});

const answer = whisperResp.result;
if (answer.status === "answered") {
  await pushToKB({
    question_class: "parental-leave-adoption",
    policy_ref: "hr.parental-leave.v3#adoption-eligibility",
    answer: answer.choice
  });
}
\end{lstlisting}

The answer is recorded as a workspace policy reference. Next time the question arises, the bot quotes the policy. The thirty-second tap is the protocol's contribution to the SME team's working knowledge base.

\subsection{A pressure drop on the fill-finish line}

03:14 on a Wednesday. A predictive-maintenance agent monitoring an isolator on a fill-finish line at a GMP-regulated biopharmaceutical site flags a deviation: differential pressure on iso-3 dropped to 8.2 millibar against a 10.0 minimum, for six minutes, during batch BX-48219. What happens next involves Annex 11 (electronic systems), Annex 1 (sterile manufacture), ICH Q9 (causation), ICH Q10 (management review evidence), and when the inspector arrives, possibly six months later, the question of whether the batch should have been released, by whom, on what evidence, under which procedure.

Today this story lives in four systems. The historian (typically AVEVA PI) has the pressure trace; the electronic quality management system (Veeva Vault, MasterControl, or equivalent) has the deviation record; the site's lean daily management board has the shift leader's note; the batch record is somewhere between the ERP and a printed paper file.

A \chap{}-instrumented site puts the whole story in one chain. The agent's flag:

\begin{lstlisting}[caption={Predictive-maintenance agent deviation flag.},label={lst:case10-flag}]
{
  "method": "task.create",
  "params": {
    "workspace": "wsp_site_fill_finish_b3",
    "from": "agent:pdm-isolator-monitor#v2.4",
    "kind": "deviation_review",
    "artefact": {
      "kind": "deviation_flag",
      "logical_id": "lgl_dev_2026_05_17_B3_001",
      "subject": "Isolator iso-3 dP below threshold during batch BX-48219",
      "evidence": {
        "historian_tag": "PI:ISO3.DP",
        "window": ["2026-05-17T03:14:00Z", "2026-05-17T03:20:00Z"],
        "value_min": 8.2,
        "threshold": 10.0
      }
    },
    "routing_hints": {
      "criticality": "high",
      "risk_tier": "GMP-batch-affecting"
    },
    "policy_refs": ["GMP.Annex-1.s.4", "site-procedure.fill-finish.v12"]
  }
}
\end{lstlisting}

Eight hours later, after investigation, the Qualified Person signs the release decision:

\begin{lstlisting}[caption={Qualified Person release.},label={lst:case10-release}]
{
  "method": "decide.approve",
  "params": {
    "task_id": "tsk_qp_release_BX_48219",
    "from": "human:qp.tanaka@biopharma.com",
    "logical_id": "lgl_batch_BX_48219_disposition",
    "signature_meaning": "qp_release",
    "rationale": "Deviation investigated; root cause = transient HVAC pressure imbalance; product not affected; CAPA-2026-0419 raised for HVAC tuning. Batch released under Annex 1 s.4.30.",
    "policy_refs": ["GMP.Annex-1.s.4.30"]
  }
}
\end{lstlisting}

When the inspector visits in October, the compliance officer's preparation is a single query. The Python client makes this concise:

\begin{lstlisting}[language=Python,caption={Compliance officer prepares for inspection.},label={lst:case10-prep}]
import chap

ws = chap.connect(
    "https://coordinator.site.example/chap",
    workspace="wsp_site_fill_finish_b3"
)

chain = ws.audit_read(
    logical_id="lgl_batch_BX_48219_disposition",
    include_referenced=True
)

for entry in chain:
    print(entry.ts, entry.method, entry.from_, "->", entry.summary)
    if (entry.method == "decide.approve"
        and entry.signature_meaning == "qp_release"):
        print(f"  QP credential: {entry.identity.credential_id}")
        print(f"  SCITT receipt: {entry.scitt_receipt_url}")
\end{lstlisting}

Two profiles do the heavy lifting here. \texttt{identity-vc/1.0} carries the QP's regulatory status as a verifiable credential rather than a string the auditor has to trust the site to render correctly. \texttt{audit-scitt/1.0} anchors the chain to a transparency log so the inspector verifies integrity without trusting the site's systems. The rest (\texttt{review/1.0} for the deviation review, \texttt{modes/1.0} for the PdM agent's documented promotion, \texttt{security-signed/1.0} for Annex 11 signature parity, \texttt{handoff/1.0} for the shift handover that carried batch context) compose around the QP signature as the load-bearing event.

\subsection{Motor claims and the bereavement that changes everything}

A UK motor insurer, regulated by the Financial Conduct Authority and subject to the Consumer Duty, runs an AI agent that handles first-notification-of-loss intake, drafts coverage assessments, and recommends settlements. Humans approve everything customer-facing.

A claimant calls about a side-impact incident the previous Saturday. During the call the claimant mentions, in passing, that they were driving to their mother's funeral when the collision happened. The agent's intake module flags the disclosure as a vulnerability indicator.

\begin{lstlisting}[caption={Vulnerability-aware claim triage.},label={lst:case11-fnol}]
{
  "method": "task.create",
  "params": {
    "workspace": "wsp_motor_claims_uk",
    "kind": "fnol_triage",
    "from": "agent:fnol-triage-bot#v3.2",
    "input": {
      "claim_ref": "MTR-2026-198421",
      "vulnerability_flags": ["recent_bereavement_in_household"]
    },
    "routing_hints": {
      "criticality": "high",
      "risk_tier": "consumer-duty-vulnerable",
      "max_review_lapse_hours": 4
    },
    "policy_refs": ["FCA.PS22-9", "internal.vulnerability-policy.v4"]
  }
}
\end{lstlisting}

The vulnerability flag forces senior-handler routing within four hours regardless of claim size. A senior handler picks up the file, adjusts the settlement to remove pre-existing damage the AI mis-classified, and applies the firm's vulnerability policy:

\begin{lstlisting}[caption={Senior-handler override applying vulnerability policy.},label={lst:case11-override}]
{
  "method": "decide.override",
  "params": {
    "task_id": "tsk_settlement_MTR-2026-198421",
    "from": "human:harper.s@insurer.com",
    "logical_id": "lgl_claim_MTR-2026-198421",
    "intent_preserved": true,
    "diff": [
      { "op": "replace", "path": "/settlement/amount", "from": 4200, "to": 4800 },
      { "op": "add", "path": "/settlement/goodwill",
        "value": { "amount": 200, "reason": "bereavement_acknowledgement" } }
    ],
    "rationale": "Customer disclosed recent bereavement; vulnerability policy applied. Rear bumper scratches the AI flagged appear to be pre-existing, common on this model. Net adjustment of 600.",
    "tags": ["vulnerability-policy-applied", "ai-damage-misclassified", "fair-outcome-adjustment"],
    "policy_refs": ["FCA.PS22-9.fair-outcome", "internal.vulnerability-policy.v4#bereavement"]
  }
}
\end{lstlisting}

A year later, an FCA supervision visit. The firm's vulnerable-customer-outcomes file is one of the routine asks. The team prepares its evidence by replaying the chain:

\begin{lstlisting}[language=Python,caption={Supervision evidence preparation.},label={lst:case11-evidence}]
chain = ws.audit_read(logical_id="lgl_claim_MTR-2026-198421")

trigger = next(e for e in chain if e.method == "task.create")
review = next(e for e in chain if e.method == "decide.override")
print(f"Vulnerability flag at: {trigger.ts}")
print(f"Senior handler decision at: {review.ts}")
print(f"Lapse: {review.ts - trigger.ts}")
print(f"Handler: {review.from_} (credential: {review.identity.credential_id})")
print(f"Policy applied: {review.policy_refs}")
print(f"Rationale: {review.rationale}")
\end{lstlisting}

The handler's name and credential are on the envelope. Under personal-accountability regimes, this is the kind of evidence that protects the named individual as well as the firm.

\subsection{Five years on, the regulator asks}

A wealth-management firm of around two hundred advisors operates under whichever personal-accountability regime applies (SOX-404 for US-listed firms, the IAF/SEAR framework in Ireland, the Senior Managers and Certification Regime in the United Kingdom). Advisors use an AI assistant to draft suitability assessments, projections, and rebalancing recommendations.

An advisor leaves the firm. Three months after she leaves, a regulator opens a thematic inspection covering advice given to fifty clients across two years, fifteen of which were hers. The question the inspection asks, in regulatory language, is whether the firm's controls operated effectively \emph{throughout} the period, not just when last sampled.

Every piece of advice is a task; the AI's proposal is an artefact; the advisor's approval is signed with her OIDC-bound key, references the version of the firm's advice rubric in effect at the time, and carries her regulatory credential.

\begin{lstlisting}[caption={Signed advice approval with rubric version captured in policy refs.},label={lst:case12-advice}]
{
  "method": "decide.approve",
  "params": {
    "task_id": "tsk_rebalance_advice_client_4821",
    "from": "human:advisor.lee@firm.com",
    "logical_id": "lgl_client_4821_advice_q2_2024",
    "based_on_artefact_id": "art_ai_rebalance_proposal_...",
    "signature_meaning": "advice_authorised",
    "rationale": "Client in accumulation phase, risk profile last assessed Nov 2023, no material change. AI proposal accepted with one adjustment: emerging-markets weight reduced from 12 per cent to 9 per cent per client's explicit Q1 instruction.",
    "policy_refs": ["firm.advice-policy.v8", "firm.advice-rubric.v3.2", "MiFID-II.suitability"]
  }
}
\end{lstlisting}

The advisor's CF1 status is carried separately by \texttt{identity-vc/1.0} as a verifiable credential whose issuer is the regulator. The credential binds to the participant identity; it is not a magic field on the override.

Five years later, the inspection becomes a query rather than a forensic reconstruction:

\begin{lstlisting}[language=Python,caption={Control-reliance evidence across the inspection window.},label={lst:case12-query}]
from datetime import datetime

audit = ws.audit_read(
    filter={
        "method": "decide.approve",
        "from": "human:advisor.lee@firm.com",
        "signature_meaning": "advice_authorised",
        "since": datetime(2021, 1, 1),
        "until": datetime(2026, 1, 1)
    }
)

rubric_versions = {}
for e in audit:
    rubric = next((p for p in e.policy_refs if "advice-rubric" in p), None)
    rubric_versions.setdefault(rubric, []).append(e)

for v, entries in sorted(rubric_versions.items()):
    first = min(e.ts for e in entries)
    last = max(e.ts for e in entries)
    print(f"{v}: {len(entries)} approvals, active {first} to {last}")
\end{lstlisting}

The verifiable-credentials profile carries the advisor's regulatory status as a credential whose issuer is the regulator itself, not the firm. The audit-SCITT profile anchors the chain to an external transparency service, which means the inspector verifies chain integrity without trusting the firm's own systems. The modes profile shows the AI assistant's promotion history.

\subsection*{What is happening here}

Read these twelve in sequence and a small set of mechanisms keeps surfacing in different costumes. A typed override that carries the diff and the reason. A clarifying question with a default if nobody answers. A handoff that preserves context across a shift. A signed approval someone outside the system can verify. The pattern is not novel: every team in these scenarios already does these things in chat threads, in ticket comments, in email subject lines, in spreadsheets. What they don't have is a common envelope to put them in.

The larger cases are not technically harder than the smaller ones. The fill-finish line and the wealth-management firm run on the same primitives as the solo developer. They simply run more of them, with stronger signatures, against longer retention windows, for more demanding readers. The smallest useful adoption begins with whichever situation looks closest to the reader's own week.

\end{document}